\documentclass[journal, onecolumn]{IEEEtran}
\IEEEoverridecommandlockouts
\usepackage[normalem]{ulem}
\usepackage[english]{babel}
\usepackage[utf8]{inputenc}
\usepackage{wrapfig, blindtext}
\usepackage{arydshln}

\usepackage{float}
\usepackage{algorithm}
\usepackage{algorithmic}
\usepackage{url}
\usepackage[dvipsnames]{xcolor}
\usepackage{amsfonts}
\usepackage{amssymb,amsmath}

\usepackage{booktabs}
\usepackage{graphicx}
\usepackage{caption}
\usepackage{cases}
\usepackage{comment}
\usepackage{multirow}
\usepackage{tikz}

\usepackage{epstopdf}
\usepackage{hyperref}
\hypersetup{
    colorlinks=true,
    citecolor=blue,  
    linkcolor=black, 
    urlcolor=blue
}

\usepackage[caption=false, skip=0pt, font=scriptsize]{subfig}
\usepackage{lipsum}
\usepackage{xargs}
\usepackage{gensymb}
\usepackage{mathtools,extarrows}

\usepackage[authoryear]{natbib}

\newcommand{\hg}[1]{\colorbox{green}{#1}}
\newcommand{\hy}[1]{\colorbox{yellow}{#1}}
\newcommand{\ho}[1]{\colorbox{orange}{#1}}

\usepackage{tikz}
\usepackage{pgfplots}
 \pgfplotsset{compat=1.18}
 \usepgfplotslibrary{statistics}
\usepackage{amsmath}

\usetikzlibrary{
  arrows.meta,                        
  backgrounds,                        
  positioning, fit,
  ext.paths.ortho,                    
  ext.positioning-plus,               
  ext.node-families.shapes.geometric, 
  calc} 
\title{\bf 
 \Large

M$^3$RS: Multi-robot, Multi-objective, and Multi-mode Routing and Scheduling}

\author{
    Ishaan Mehta$^{a}$,  
    Junseo Kim$^{b}$,  
    Sharareh Taghipour$^{a}$,  
    Sajad Saeedi$^{c}$  
    \\
    $^{a}$ Toronto Metropolitan University, Toronto, ON, Canada \\
    $^{b}$ Delft University of Technology, Delft, Netherlands \\
    $^{c}$ University College London, London, UK \\
    {\small Email: ishaan.mehta@torontomu.ca, j.kim-18@student.tudelft.nl, sharareh@torontomu.ca, s.saeedi@ucl.ac.uk}
}

\begin{document}
\maketitle 
\vspace{-5 mm}
\begin{abstract}
\textcolor{black}{Task execution quality significantly impacts multi-robot missions, yet existing task allocation frameworks rarely consider quality of service as a decision variable, despite its importance in applications like robotic disinfection and cleaning. We introduce the multi-robot, multi-objective, and multi-mode routing and scheduling (M$^3$RS) problem, designed for time-constrained missions. In M$^3$RS, each task offers multiple execution modes with varying resource needs, durations, and quality levels, allowing trade-offs across mission objectives. M$^3$RS is modeled as a mixed-integer linear programming (MIP) problem and optimizes task sequencing and execution modes for each agent. We apply M$^3$RS to multi-robot disinfection in healthcare and public spaces, optimizing disinfection quality and task completion rates. Through synthetic case studies, M$^3$RS demonstrates 3\%–46\% performance improvements over the standard task allocation method across various metrics. Further, to improve compute time, we propose a clustering-based column generation algorithm that achieves solutions comparable to or better than the baseline MIP solver while reducing computation time by 60$\%$. We also conduct case studies with simulated and real robots. Experimental videos are available on the project page\footnote{\href{https://sites.google.com/view/g-robot/m3rs/}{https://sites.google.com/view/g-robot/m3rs/}}.}
\end{abstract}
\hspace{-1 mm}
\vspace{-5 mm}

\section{Introduction and Related Works}
\par

\noindent \textcolor{black}{Multi-robot teams are widely used in diverse applications such as search and rescue~\citep{liu2016multirobot,olson2012progress}, environment mapping~\citep{saeedi2016multiple, lajoie2022towards}, disaster management response~\citep{phan2008cooperative}, marine operations~\citep{thompson2019review}, and agricultural tasks~\citep{das2025unified}. In healthcare settings, these teams can play a vital role in maintaining hygiene to prevent hospital-acquired infections.} 
Disinfection robots offer an effective solution by automating surface disinfection, adding an extra layer of protection. A team of these robots can enhance both spatial and temporal efficiency in maintaining hygiene standards in hospitals~\citep{mehta2023uv}.   Coordinating a fleet of disinfection robots requires making two key decisions:  i)~assigning tasks to robots, and ii)~determining the level of disinfection quality for each task.  Although higher disinfection quality is preferable, it also requires more time and resources, which may not always be feasible. Task allocation for multi-robot systems has been well-studied in the literature~\citep{korsah2013comprehensive}, but the decision regarding quality of service (QoS) for individual tasks remains largely unexplored. In disinfection applications, QoS is critical, as some surfaces, such as those in patient rooms and ICUs, require higher hygiene standards. This work addresses the challenge of selecting tasks and determining the appropriate disinfection quality/mode for each task.

\par
\textcolor{black}{Multi-Robot Task Allocation (MRTA) algorithms~\citep{gerkey2004formal, korsah2013comprehensive} assign tasks to robots to optimize a joint mission objective. A key factor in evaluating these allocations is the QoS of the mission. \textcolor{black}{The target of QoS} varies depending on the application. \textcolor{black}{For instance, QoS can refer to the quality of communication between robots and a base station~\citep{khan2022emerging, wang2023development}, reduced task completion delays~\citep{botros2023optimizing}, quantity of area coverage for agricultural survey applications~\citep{barrientos2011aerial},} reduced average customer waiting time~\citep{miller2017predictive}, or optimal worker productivity and reduced fatigue~\citep{calzavara2023multi}. In many cases, QoS competes with other mission objectives, requiring trade-offs among different priorities~\citep{hayat2020multi, flushing2017simultaneous, calzavara2023multi,liang2023bi}.} \textcolor{black}{While QoS has been considered to some extent in the MRTA literature, no prior work incorporates QoS as a decision variable specifically for disinfection applications. Introducing QoS as an explicit decision variable enables more precise control over the trade-offs between disinfection quality and other mission objectives. In disinfection scenarios, for example, it allows for optimizing hygiene standards while maximizing task completion within the available mission time.
}



\begin{figure}[htbp]
\centering
\begin{tikzpicture}[
    >={Latex[length=3pt]},
    dashedarrow/.style = {-Latex, thick, dashed},
    agent/.style      = {regular polygon, regular polygon sides=3,
                         minimum size=12mm, inner sep=0pt, draw, thick},
    task/.style       = {rectangle, draw, thick, minimum width=18mm,
                         minimum height=8mm, font=\footnotesize, align=center},
    mode1/.style      = {task, fill=green!30},
    mode2/.style      = {task, fill=yellow!30},
    mode3/.style      = {task, fill=red!20},
    missed/.style     = {task, fill=gray!35},
    lab/.style        = {font=\scriptsize},
]

\coordinate (shiftA) at (-3,  0);   
\coordinate (shiftB) at (5,  0);   
\coordinate (shiftC) at (-0.5, -5);   

\begin{scope}[shift={(shiftA)}]
    \node[agent, fill=red!60]             (A1) at (-2, 1.6) {};
    \node[agent, fill=teal!60]            (A2) at (-2,-0.6) {};
    \node[lab, anchor=east]               at ($(A1.west)+(-0.4,0)$) {Agent 1};
    \node[lab, anchor=east]               at ($(A2.west)+(-0.4,0)$) {Agent 2};

    \node[mode1]  (T1) at (0, 1.6) {Task 1};
    \node[missed] (T2) at (2, 1.6) {Task 2};

    \node[mode1]  (T4) at (0,-0.6) {Task 4};
    \node[missed] (T3) at (2,-0.6) {Task 3};

    \draw[dashedarrow] (A1) -- (T1);
    \draw[dashedarrow] (T1.south) to[out=-90, in=-90] (A1.south);
    \draw[dashedarrow] (A2) -- (T4);
    \draw[dashedarrow] (T4.south) to[out=-90, in=-90] (A2.south);

    \node[lab] at ($(A2.south)!0.5!(T3.south) + (0,-0.6)$) {(a)};
\end{scope}

\begin{scope}[shift={(shiftB)}]
    \node[agent, fill=red!60]  (B1) at (-2, 1.6) {};
    \node[agent, fill=teal!60] (B2) at (-2,-0.6) {};
    \node[lab, anchor=east]    at ($(B1.west)+(-0.4,0)$) {Agent 1};
    \node[lab, anchor=east]    at ($(B2.west)+(-0.4,0)$) {Agent 2};

    \node[mode3] (B1T1) at (0, 1.6) {Task 1};
    \node[mode3] (B1T2) at (2.5, 1.2) {Task 2};

    \node[mode3] (B2T4) at (0,-0.6) {Task 4};
    \node[mode3] (B2T3) at (2.5,-0.8) {Task 3};

    \draw[dashedarrow] (B1)   -- (B1T1);
    \draw[dashedarrow] (B1T1) -- (B1T2);
    \draw[dashedarrow] (B1T2.south) to[out=-40, in=-20] (B1.south);
    \draw[dashedarrow] (B2)   -- (B2T4);
    \draw[dashedarrow] (B2T4) -- (B2T3);
    \draw[dashedarrow] (B2T3.south) to[out=181, in=-40] (B2.south);

    \node[lab] at ($(B2.south)!0.5!(B2T3.south) + (0,-0.6)$) {(b)};
\end{scope}

\begin{scope}[shift={(shiftC)}]
    \node[agent, fill=red!60]  (C1) at (-2.5, 1.25) {};
    \node[lab, anchor=east]    at ($(C1.west)+(-0.4,0)$) {Agent 1};

    \node[agent, fill=teal!60] (C2) at (4.2, 1.25) {};
    \node[lab, anchor=east]    at ($(C2.west)+(-0.1,0.0)$) {Agent 2};

    \node[mode2] (C1T1) at (0,  1.25) {Task 1};
    \node[mode2] (C1T2) at (-1.0, -0.15) {Task 2};

    \node[mode1] (C2T4) at (6.2, 1.7) {Task 4};
    \node[mode3] (C2T3) at (5.4, 0.0) {Task 3};

    \draw[dashedarrow] (C1)   -- (C1T1);
    \draw[dashedarrow] (C1T1) -- (C1T2);
    \draw[dashedarrow] (C1T2) -- (C1);   

    \draw[dashedarrow] (C2)   -- (C2T4);
    \draw[dashedarrow] (C2T4) -- (C2T3);
    \draw[dashedarrow] (C2T3) -- (C2);

    \node[lab] at ($(C1.south west)!0.5!(C2.south east)+(1.5,-0.9)$) {(c)};
\end{scope}

\draw[dotted, thick] ($(shiftA)+(4,2.2)$) -- ($(shiftA)+(4,-2.2)$);  
\draw[dotted, thick] ($(shiftA)+(-2.8,-2.2)$) -- ($(shiftA)+(11.8,-2.2)$); 

\begin{scope}[xshift=-2.5cm,yshift=-6.8cm]
    \node[mode1, minimum width=0.4cm, minimum height=0.3cm, label={[lab]right:Mode 1}] (L1) {};
    \node[mode2, minimum width=0.4cm, minimum height=0.3cm, right=1.5cm of L1, label={[lab]right:Mode 2}] (L2) {};
    \node[mode3, minimum width=0.4cm, minimum height=0.3cm, right=1.5cm of L2, label={[lab]right:Mode 3}] (L3) {};
    \node[missed, minimum width=0.4cm, minimum height=0.3cm, right=1.5cm of L3, label={[lab]right:Missed Task}] (L4) {};
\end{scope}
\end{tikzpicture}
\caption{Traditional Multi-Robot Task Allocation models overlook Quality of Service (QoS) as a decision variable, assuming agents provide a uniform service level, which can lead to inefficiencies in routing and scheduling under time and resource constraints. The visualization compares three cases: (a)~Agents operate at the highest service level (Mode 1), ensuring maximum quality but consuming more time and energy, resulting in missed tasks (e.g., Tasks 2 and 3). (b)~Agents operate at the lowest service level (Mode 3), completing all tasks but at reduced quality. (c)~M$^3$RS incorporates QoS as a decision variable, allowing agents to balance quality and quantity dynamically--choosing from Mode 1, Mode 2 (moderate quality), and Mode 3--ensuring all tasks are addressed at appropriate service levels. Unlike conventional methods, M$^3$RS enables flexible task allocation by integrating QoS into decision-making, making it well-suited for multi-robot applications such as disinfection and cleaning, where balancing quality and quantity is crucial.}
\label{fig:fig1_vis}
\end{figure}

\textcolor{black}{Additional decision variables for QoS can further increase the computation complexity of the MRTA problem, which is inherently $\mathcal{NP}$-hard. While heuristics and metaheuristics can provide faster solutions, they often compromise optimality~\citep{braekers2016vehicle}. Learning-based approaches require extensive training data and may struggle to generalize~\citep{bogyrbayeva2022learning}.} \textcolor{black}{To efficiently handle the added complexity of task-level QoS variables, we propose a clustering-based column generation algorithm. Column generation \citep{desrosiers2005primer} iteratively refines solutions by solving smaller subproblems, allowing for faster convergence to high-quality solutions. The integration of clustering further helps in reducing the search space to find good quality solutions.}
\par
To the best of our knowledge, this work presents the first \textcolor{black}{MRTA problem}  to incorporate task-level QoS using multiple execution modes for quality-dependent multi-robot missions. We introduce the Multi-Robot, Multi-Objective, and Multi-Mode Routing and Scheduling (M$^3$RS) problem, which enables adjustment of task-level QoS to optimize trade-offs among mission objectives. M$^3$RS is particularly relevant for applications where tasks are spatially distributed, have varying service and resource requirements, and involve multiple mission criteria. Each robot can execute tasks in different modes, where each mode dictates task quality, resource consumption, and service time.
A visual representation of \textcolor{black}{M$^3$RS compared to the typical MRTA problem is provided in Fig.~\ref{fig:fig1_vis}}. Traditional MRTA models ignore QoS, while we consider it as a decision variable.  
\textcolor{black}{We use M$^{3}$RS for multi-robot disinfection, where both disinfection quality and task throughput are critical. High-quality disinfection, however, increases time and energy costs, limiting the number of tasks performed. M$^{3}$RS addresses this trade-off by leveraging task-execution modes to balance quality and efficiency.} Our contributions are briefly summarized below:
\begin{enumerate}
    \item \textcolor{black}{We propose a novel MRTA problem called Multi-Robot, Multi-Objective, and Multi-Mode Routing and Scheduling (M$^3$RS) that explicitly considers task-level QoS.}
    \item \textcolor{black}{We apply M$^3$RS to a multi-robot disinfection scenario. Case studies on synthetic data show that M$^3$RS solutions outperform solutions of standard MRTA and routing formulations with a margin ranging from 3$\%$-46$\%$ across multiple metrics pertaining to quality of disinfection and quantity of tasks attempted.}  
    \item \textcolor{black}{We propose clustering-based column generation (CCG) for solving M$^{3}$RS. CCG generates competitive solutions with compute time reduced by 60$\%$.}
    \item \textcolor{black}{We present simulated and real-world experiments for multi-robot disinfection missions.}
\end{enumerate}

This paper is organized as follows: The general formulation and optimization of M$^3$RS is given in Section~\ref{sec:m3rs}. The results of the synthetic, simulated, and real-world experiments are presented in Section~\ref{sec:exp}. Finally, Section~\ref{sec:conc} includes future works and conclusions.

\section{Literature Review: QoS in Task Assignment}\label{sec:lit}
\textcolor{black}{QoS has also received attention in the literature on routing and production scheduling. \citet{exposito2019quality} considered the timeliness of vehicles as a QoS indicator for vehicle routing. Similarly, QoS has been studied for home healthcare applications, where it involves routing nurses to different patients for healthcare-related services, depending on the patient receiving care from a nurse with the desired experience \citep{kordi2023multi, di2021routing}. In supply chain and production planning, QoS has been considered as the existence of minimum stock \citep{anli2007tractable}. In logistics application, customer satisfaction~\citep{liang2023bi} and optimizing the service times~\citep{gonzalez2024bi} have also been considered as indicators of QoS. Most of these existing works modify the task allocation to achieve different trade-offs between QoS and other objectives. However, there are no works that focus on the deliberate degradation or improvement of QoS offered to an individual task as a means to achieve an overall trade-off for various mission objectives.}
\par 
\textcolor{black}{Task execution modes refer to distinct operational settings for completing tasks, often chosen based on environmental constraints or desired performance metrics such as tardiness and resource consumption. This concept has been extensively studied in domains like project scheduling and vehicle routing. In project scheduling, execution modes typically represent alternative ways to perform an activity—for instance, allocating more resources to reduce its duration \citep{hartmann2022updated,wkeglarz2011project,tiwari2009scheduling}. In routing applications, mode selection has been used to choose between transportation methods for healthcare service providers \citep{braekers2016bi}, manage hybrid vehicle driving strategies to minimize emissions and costs \citep{seyfi2022multi}, and optimize battery recharging strategies in electric vehicle fleets \citep{zhang2024novel}. In MRTA, the use of task execution modes has been limited. Some studies have explored their use to determine whether human-robot collaboration is beneficial for improving time efficiency \citep{liau2022genetic, ferreira2021scheduling, parreno2024solving}. However, the use of task execution modes to explicitly control the QoS provided to individual tasks remains largely unexplored in robotic applications. This is particularly important in domains such as multi-robot disinfection, where higher disinfection quality (achieved through higher dosages, for example), has clear physical significance but also incurs greater time and energy costs, potentially reducing overall task throughput. We use task execution modes in M$^3$RS to explore such trade-offs in the context of MRTA.}

\section{Multi-Robot Multi-Objective Multi-Mode Routing and Scheduling}\label{sec:m3rs}

\textcolor{black}{In this section, first, the general optimization model for M$^3$RS problem is presented. Then, we present a column generation (CG) scheme to solve the optimization model with faster computation times. Our novel contribution is the use of clustering heuristics, which helps solve the problem more quickly and find solutions with better objective values. Finally, the formulation is applied to the multi-robot disinfection application.}

\begin{table}[t]
    \normalsize
    \centering
    \caption{Notation for M$^3$RS. 
    }
    \begin{tabular}{|c|c|p{10.5cm}|p{4cm}|}
         \hline
         No & Notation & Description & Comment \\
         \hline
         1 & $\mathcal{A}$  & Set of agents. & \multirow[t]{3}{*}{\parbox{4cm}{Problem Sets 
         }}\\
         2 & $\mathcal{T}$  & Set of tasks. &  \\
         3 & $\mathcal{M}_i$ &  Set of modes for task $i$, where $\mathcal{M}_i = \{1,2 ... M\}$. &   \\
         \hline
         4 & $R$ & Number of objective criteria.  & \multirow[t]{3}{*}{\parbox{4cm}{Objective Function 
         }} \\
         5 & $f_r$ & $r^{th}$ objective criterion. & \\
         6 & $\lambda_r$ & The scalar preference for  $r^{th}$ objective Function. & \\
         \hline
         7 & $q_{i\scriptsize{\xrightarrow{}}j}$ & Energy consumed for traveling from task $i$ to task $j$. & \multirow[t]{7}{*}{\parbox{4cm}{Energy Constraints 
         }} \\
         8 & $q_{\text{idle}}^k$ & Energy consumption during idling by agent $k$. &\\
         9 & $q_{i,m}$ & Energy consumption of task $i$ in mode $m$. & \\
         10 & $q_{\text{travel}}$ & Average current consumption in traveling at a constant speed. &\\
         11 & $q_{\text{tc}}$ & Average current consumption while completing tasks.& \\
         12 & $q_{\text{idling}}$ & Average current consumption while idling. &\\
         13 & $Q$ & Total \textcolor{black}{energy} capacity of an agent. &\\         
         \hline
         14 & $\sigma_{i,m}$ & Time  to complete task $i$ in mode $m$. & \multirow[t]{4}{*}{\parbox{4cm}{Time Constraints  
         }}  \\
         15 & $[a_i, b_i]$ & Start and end time for task $i$. &\\
         16 & $T_H$ & Mission time or time available to complete all tasks.& \\
         17 & $t_{i,j}$ & Travel time from task $i$ to task $j$. & \\
         \hline         
         18 & $p_{i,m}$ &   The quality reward of doing task $i$ in mode $m$. & \multirow[t]{3}{*}{\parbox{4cm}{Disinfection Problem 
         }}\\
         19 & $p^{max}$ & The reward for the highest possible quality of disinfection. & \\
         20 & $\eta$ & Load factor: \textcolor{black}{mission workload relative to the robot’s available capacity.}&\\
         \hline
         21 & $x^{k}_{i{\xrightarrow{}}j}$ & A binary decision variable for if agent $k$ goes from task $i$ to task $j$.& \multirow[t]{3}{*}{\parbox{4cm}{Decision Variables 
         }} \\
         22 & $y^{k}_{i,m}$ & A binary decision variable for task  $i$ done in mode $m$ by agent $k$.& \\
         23 & $s^{k}_i$ & A decision variable for arrival time of agent $k$ to task $i$.& \\        
         \hline     
    \end{tabular}
    \vspace{-0.2in}
    \label{tab:nots}
\end{table}
\vspace{-0.04in}
\subsection{Optimization Model}

\textcolor{black}{A solution for the M$^3$RS problem determines routes, schedules, and task execution modes for a set of agents $\mathcal{A}$, and a set of tasks $\mathcal{T}$, distributed across a physical space.} \textcolor{black}{Refer to Table~\ref{tab:nots} for notation and symbols used to formulate the problem. Here, all agents are identical, and only one agent can service each task. We use  $i$ and $j$  to denote different tasks in $\mathcal{T}$ and $k$ to denote $k^{th}$ agent in the rest of the notation. The time window for a task $i$ specifies a starting time, $a_i$, and finishing time, $b_i$, within which the task must be serviced.  The entire multi-robot operation must be done in a fixed mission time of $T_H$ hours. Further, each task $i$ has an associated mode set $\mathcal{M}_i = \{1,2 ... M\}$, representing available task service modes.
Each task can have distinct modes of service available, so each task has a distinct mode set. An agent services the task in one of the modes available in the associated mode set. The selected mode $m \in \mathcal{M}_i$ for task $i$ has a corresponding service time ($\sigma_{i,m}$) and energy requirement ($q_{i,m}$) from the agent. Performing a task in a mode with higher service time and resource consumption would lead to lower availability of agents for other tasks. The overall scheduling and allocation mission is optimized for $R$ objective functions. }

\par
The formulation for M$^3$RS is expressed as a mixed integer with the following decision variables: 
\begin{enumerate}
    \item The binary decision variable $x^{k}_{i{\xrightarrow{}}j}\in \{0,1\}$, for assignment and transition of agent $k$ from task $i$ to $j$; 
    \item The binary decision variable $y^{k}_{i,m}\in  \{0,1\}$, for allocation of mode $m$ for task $i$ when executed by agent $k$; and 
    \item The real continuous variable $s^{k}_i \in \mathbb{R}$ denotes arrival time of agent $k$ to task $i$.
\end{enumerate}

With these descriptions, the optimization problem is expressed as follows, optimizing the variables mentioned above:
%
\small
\begin{equation} \label{eq:sched1}
    \underset{x^{k}_{i{\xrightarrow{}}j}, y^{k}_{i,m},s^{k}_i}{\max} \, \sum_{r=0}^{R} \sum_{k \in \mathcal{A}} \sum_{i \in \mathcal{T}}  \sum_{j \in \mathcal{T}} \sum_{m \in \mathcal{M}_i} \lambda_r \cdot f_{r}(x^{k}_{i{\xrightarrow{}}j}, y^{k}_{i,m}, s^{k}_i),
\end{equation}
\normalsize
\noindent subject to the constraints in Eq.~\eqref{eq:scstr1}-\eqref{eq:scstr10b}. 

Eq.~\eqref{eq:sched1} is the scalar objective function that maximizes the convex combination of $R$ objective criteria. Here, $f_r$ is $r^{th}$ objective criteria for $r = 1,2,...R$.  The user-specified parameter $\lambda_r$ represents the trade-off preference for $r^{th}$ objective criterion. Each objective criterion can depend on task allocation, selected mode, and arrival times.

The constraints of the optimization, described through Eq.~\eqref{eq:scstr1}-\eqref{eq:scstr10b}, are categorized into five categories, as follows:
\begin{enumerate}
\item Task assignment constraint, Eq.~\eqref{eq:scstr1},  
\item Feasibility constraints, Eq.~\eqref{eq:scstr1b}-\eqref{eq:scstr4b}, 
\item Energy consumption constraint, Eq.~\eqref{eq:scstr5}, 
\item Time constraints, and Eq.~\eqref{eq:scstr6}-\eqref{eq:scstr7a}, and 
\item Domain and preference constraints, Eq.~\eqref{eq:scstr8}-\eqref{eq:scstr10b}. 
\end{enumerate}
These constraints are described next. 
\par
\subsubsection{Task Assignment Constraint}
Given by Eq.~\eqref{eq:scstr1}, the task assignment constraint ensures that each task is done at most once:
\small
\begin{equation} \label{eq:scstr1}
\sum_{k \in \mathcal{A}} \sum_{j \in \mathcal{T}} x^{k}_{i{\xrightarrow{}}j} \, \leq \, 1 \,\,\, \forall i \in \mathcal{T} \backslash 0. 
\end{equation}
\normalsize

\noindent Here, $i=0$ is excluded ($\mathcal{T}\backslash0$) as it corresponds to the home depot. This is done to maintain feasibility, as all agents must leave the home depot. There might be missions where some tasks must be completed. For such cases, where the tasks must be completed, the inequality can be replaced by equality.  

\subsubsection{Feasibility Constraints} 
Given by Eq.~\eqref{eq:scstr1b}-\eqref{eq:scstr4b}, the feasibility constraints ensure the solutions are logical.

\small
\begin{equation} \label{eq:scstr1b}
x^{k}_{i{\xrightarrow{}}i} \, = \, 0  \,\, \forall k \in \mathcal{A} \,\, \forall i \in \mathcal{T},
\end{equation}

\begin{equation} \label{eq:scstr2}
\sum_{i \in \mathcal{T}} x^{k}_{i{\xrightarrow{}}0} \, = \, 1 \,\,\, \forall k \in \mathcal{A},
\end{equation}

\begin{equation} \label{eq:scstr3}
\sum_{j \in \mathcal{T}} x^{k}_{0\scriptsize{\xrightarrow{}}j} \, = \, 1 \,\,\, \forall k \in \mathcal{A},
\end{equation} 

\begin{equation} \label{eq:scstr4}
\sum_{i \in \mathcal{T}} x^{k}_{i{\xrightarrow{}}h} - \sum_{j \in \mathcal{T}} x^{k}_{h\scriptsize{\xrightarrow{}}j} \, = \, 0 \,\,\,  \forall h \in \mathcal{T} , \,\, \forall k \in \mathcal{A},
\end{equation}

\begin{equation} \label{eq:scstr4b}
    \sum_{j \in \mathcal{T}} x^{k}_{i{\xrightarrow{}}j} - \sum_{m \in \mathcal{M}_i}  y^{k}_{i,m} \, = \,  0  \, \, \forall (k,i) \in \mathcal{A} \cup \mathcal{T}\backslash0.
\end{equation}
 \normalsize

\noindent The constraint in Eq.~\eqref{eq:scstr1b} eliminates infeasible solutions where an agent transitions between the same task. The constraints in  Eq.~\eqref{eq:scstr2}-\eqref{eq:scstr3} ensure that all agents leave the home depot and return to the home depot, respectively. The constraint in Eq.~\eqref{eq:scstr4} ensures that the agent must exit the tasks they visit. \textcolor{black}{The constraint in Eq.~\eqref{eq:scstr4b}
enforces that each task $i$ can be performed in only one of the modes in $\mathcal{M}_i$. Thus, for any task $i$ selected by agent $k$, only one execution mode is used}.

\subsubsection{Energy Consumption Constraint}

The energy capacity for each agent is given by $Q$ Ah (Ampere-hours). The energy consumption for each agent is divided into performing tasks, traveling, and idling \citep{zhang2023energy}. A fixed-rate battery model is used to calculate the energy consumption, where energy consumed is proportional to idling time, task execution time, and traveling time, respectively \citep{zhang2023energy}.  Different aspects of energy expenditure can be calculated using: 
\begin{equation}
    \begin{split}
    q_{i\scriptsize{\xrightarrow{}}j} = q_{\text{travel}} \cdot t_{i,j},~~\text{\small (i.e. traveling energy)} \\
    q_{i,m} = q_{\text{tc}} \cdot \sigma_{i,m},~~\text{\small (i.e. servicing energy)} \\
    q_{\text{idle}}^k  = q_{\text{idling}} \cdot ( M_k - W_k ).~~\text{\small (i.e. idling energy)}
    \end{split}
\end{equation}
\normalsize
\noindent Here, $q_{\text{travel}}$ is the current consumption for an agent traveling at a constant speed, $q_{\text{tc}}$ is the current consumption while performing a task, and $q_{\text{idling}}$ is the current consumption for idling. $M_k$ is the total time spent by agent $k$ on the mission, and $W_k$ is the total time spent by agent $k$ traveling and executing tasks. The net energy consumed by agent $k$ in a mission is given by the LHS of Eq.~\eqref{eq:scstr5}. The relevant notation for the formulation is summarized in Table~\ref{tab:nots}.

As described by Eq.~\eqref{eq:scstr5}, the energy consumption constraint enforces that the energy expenditure of an agent in the form of traveling, doing tasks, and idling does not exceed the agent's capacity:
\begin{equation} \label{eq:scstr5}
q_{\text{idle}}^k + \sum_{i \in \mathcal{T}} (\sum_{m \in \mathcal{M}_i} q_{i,m}\cdot y^{k}_{i,m} +  \sum_{j \in \mathcal{T}}   q_{i\scriptsize{\xrightarrow{}}j} \cdot x^{k}_{i{\xrightarrow{}}j}) \, \leq \, Q \,\,\, \forall k \in \mathcal{A}.
\end{equation}

\subsubsection{Time Constraints}
Given by Eq.~\eqref{eq:scstr6}-\eqref{eq:scstr7a}, the time constraints ensure the timing of the tasks is consistent:
\small
\begin{equation}  \label{eq:scstr6}
  s^{k}_i + \sum_{m \in \mathcal{M}_i}  \sigma_{i,m} 
 \cdot y^{k}_{i,m} + t_{i,j} - s^{k}_{j} 
 \leq 0 \,\, \forall i \in \mathcal{T} \\
 \forall j \in \mathcal{T} \,\, \forall k \in \mathcal{A},
\end{equation}

\begin{equation} \label{eq:scstr7}
 a_i \leq s^{k}_i  + \sum_{m \in \mathcal{M}_i}  \sigma_{i,m} 
 \cdot y^{k}_{i,m} < b_i \,\,\, \forall i \in \mathcal{T} , \,\, \forall k \in \mathcal{A},
\end{equation}

\begin{equation} \label{eq:scstr7a}
 0 \leq s^{k}_{0} < T_H \,\, \forall k \in \mathcal{A}.
\end{equation}
\normalsize

\noindent The constraint in Eq.~\eqref{eq:scstr6} ensures that the arrival time for a subsequent task is always later than the previous task. This constraint also enables sub-tour elimination \citep{kara2004note}, where sub-tours are infeasible solutions that represent disjoint sequences of completing tasks for the same agent. The constraint in Eq.~\eqref{eq:scstr7} ensures that the agent arrives and completes the task in the selected mode within the task's prescribed time window. Finally, the constraint in  Eq.~\eqref{eq:scstr7a} ensures that all agents return to the depot within the specified time horizon $T_H$.  

\subsubsection{Domain and User Preference Constraints}
Given by Eq.~\eqref{eq:scstr8}-\eqref{eq:scstr10b}, 
the preference constraints ensure the variable and parameters follow the definitions:
\vspace{-0.001in}
\small
\begin{equation} \label{eq:scstr8}
x^{k}_{i{\xrightarrow{}}j} \in \{0,1\},
\end{equation}
\vspace{-0.1in}
\begin{equation} \label{eq:scstr9}
y^{k}_{i,m} \in \{0,1\}.
\end{equation}
\normalsize
\par 
\small
\begin{equation} \label{eq:scstr10}
0 < \lambda_r  \leq 1,
\end{equation}
\begin{equation} \label{eq:scstr10b}
\sum_{r=0}^{R} \lambda_r  = 1.
\end{equation}
\normalsize
\noindent The constraints in Eq.~\eqref{eq:scstr8}-\eqref{eq:scstr9} enforce binary values for $x^{k}_{i{\xrightarrow{}}j}$ and $y^{k}_{i,m}$, while Eq.~\eqref{eq:scstr10}-\eqref{eq:scstr10b} ensure convexity according to given user preferences.

\vspace{5mm}
\subsection{A Solution for M$^3$RS: Clustering-based Column Generation}

\textcolor{black}{M$^3$RS extends traditional routing and scheduling problems by incorporating the decision of mode selection, which affects task quality.} Routing and scheduling problems are known to be $\mathcal{NP}-$hard \citep{kumar2012survey}. Consequently, larger problem instances with many tasks and agents lead to exponential growth in the solution space and significantly burden computational resources \citep{uchoa2017new}. We use Column Generation (CG) \citep{desrosiers2005primer, lubbecke2005selected} to address the computational challenges associated with the large decision variables in the proposed formulation. The strength of CG lies in its ability to break down a complex problem into manageable subproblems while iteratively refining the solution. 

\par
\textcolor{black}{The CG algorithm decomposes a large-scale optimization problem into two components: a master problem and a subproblem (see Fig.~\ref{fig:CGnCCG}~(a)). The master problem, typically formulated as a linear program~\citep{vanderbeck2005implementing}, selects the best combination of columns (also referred to as paths or candidate solutions) from a feasible set $\Omega$. Since $\Omega$ is generally too large to consider in full, a restricted master problem is solved over a smaller subset $\Omega' \subset \Omega$. Solving this restricted problem yields a tentative solution along with dual variables. In a linear program, each constraint in the primal problem corresponds to a variable in the dual problem, known as a dual variable~\citep{bachem1992linear}. Intuitively, a dual variable represents the rate at which the optimal value of the primal objective would improve if the associated constraint were relaxed by one unit. The subproblem uses these dual variables to identify new columns that could improve the current solution. Guided by the duals, the subproblem searches $\Omega$ for high-value candidates missing from $\Omega'$. These candidates are added to $\Omega'$, and the master problem is re-solved. For example, the dual variables associated with the restricted master problem constraints in Eqs.~\eqref{eq:rmpc1}--\eqref{eq:rmpc1b} are used in the objective function of the subproblem defined in Eq.~\eqref{eq:submip}. This iterative process continues until no further columns can be found that would improve the objective value of the master problem.} 

\par
\textcolor{black}{The overall performance of the CG algorithm heavily depends on how effectively the subproblem is solved, particularly as the number of agents and tasks increases. In such cases, identifying promising solution candidates to expand the restricted set $\Omega'$ becomes computationally challenging. To address this limitation, we propose a clustering-based heuristic method, coined Clustering-based Column Generation (CCG), to solve the subproblem method efficiently. } \textcolor{black}{CCG tackles the subproblem one agent at a time, focusing on a small, representative subset of tasks, denoted by $\mathcal{T}'$ (see Fig.~\ref{fig:CGnCCG}~(b)). This subset is constructed by first clustering the original task set $\mathcal{T}$ based on task time-windows, and then uniformly sampling from these clusters to form $\mathcal{T}'$. This targeted reduction in task space allows for more scalable and efficient subproblem solving. The detailed methodology behind CCG is presented in the following section.} 
\begin{figure*}[t]
    \centering
    \subfloat[Block diagram of Column Generation (CG)]{
    \begin{tikzpicture}[
        block/.style={rectangle, draw, minimum width=2.8cm, minimum height=1cm},
        arrow/.style={->, dashed, thick},
        every node/.style={align=center}
    ]
    
    \node[block, fill=gray!20] (master) at (90:3) {Master Problem};
    \node[block, fill=lime!70] (dual) at (0:3) {Dual Values};
    \node[block, fill=gray!20] (sub) at (270:3) {Subproblem};
    \node[block, fill=yellow] (solution) at (180:3) {Subset Solution\\Pool ($\Omega' \subset \Omega$)};
    \node[block, draw=none, fill=none, below=1.2cm of solution, yshift=1.2cm] (solpool) {Solution Pool \\ ($\Omega$)};
    \begin{scope}[on background layer]
        \node[block, draw=black, dashed, fit=(solution)(solpool), fill=gray!20, inner sep=4pt] (subbox) {};
    \end{scope}

    \draw[arrow] (master.east) to[out=0, in=90] (dual.north);
    \draw[arrow] (dual.south) to[out=270, in=0] (sub.east);
    \draw[arrow] (sub.west) to[out=180, in=270] ([xshift=-1.23cm, yshift=-4pt]solution.south);
    \draw[arrow] (solution.north) to[out=90, in=180] (master.west);
    
    \end{tikzpicture}    
    }
    \subfloat[Block diagram of the proposed Clustering Column Generation (CCG)]
    {
    \begin{tikzpicture}[
        block/.style={rectangle, draw, minimum width=2.8cm, minimum height=1cm},
        smallblock/.style={rectangle, draw, fill=blue!10, minimum width=2.5cm, minimum height=0.8cm},
        arrow/.style={->, dashed, thick},
        solidarrow/.style={->, thick},
        box/.style={draw, dash dot, inner sep=0.5em, rounded corners},
    ]
    \node[block, fill=gray!20] (master) at (0,3) {Master Problem};
    \node[block, fill=yellow, align=center] (solution) at (-3,1.2) {Subset Solution\\Pool ($\Omega' \subset \Omega$)};
    \node[block, draw=none, align = center, fill=none, below=1.2cm of solution, yshift=1.2cm] (solpool) {Solution Pool \\ ($\Omega$)};
    \begin{scope}[on background layer]
        \node[block, draw=black, dashed, fit=(solution)(solpool), fill=gray!20, inner sep=4pt] (subbox) {};
    \end{scope}
    
    \node[block, fill=lime!70] (dual) at (3,1.2) {Dual Values};
    \node[smallblock] (avail) at (-1.5,-1.4) {\scriptsize Available Tasks};
    \node[smallblock] (mip) at (1.5,-1.4) {\scriptsize Create Clusters};
    \node[smallblock, align=center] (clusters) at (-1.5,-2.8) {\scriptsize Single-agent\\ \scriptsize MIP};
    \node[smallblock, align=center] (sample) at (1.5,-2.8) {\scriptsize Sampled task subset\\ \scriptsize ($\mathcal{T'}$)};
    \node[block, draw=none, fill=none] (subproblemlabel) at (0,-2.1) {Subproblem};
    \scoped[on background layer]
    \draw[arrow] (dual.south) to[out=270, in=0] (subproblemlabel.east);
    \scoped[on background layer]
    \draw[arrow] (subproblemlabel.west) to[out=180, in=270] ([xshift=-1.23cm] solution.south);
    \scoped[on background layer]
    \node[box, fit=(avail)(mip)(clusters)(sample), fill=gray!20] (subbox) {};
    \draw[arrow] (solution.north) to[out=90, in=180] (master.west);
    \draw[arrow] (master.east) to[out=0, in=90] (dual.north);
    \draw[solidarrow] (clusters) -- (avail);
    \draw[solidarrow] (avail) -- (mip);
    \draw[solidarrow] (mip) -- (sample);
    \draw[solidarrow] (sample) -- (clusters);
    \node[smallblock, draw=none, fill=blue!10, minimum width=1.8cm] (taskset) at (0,0.4) {\scriptsize Task Set ($\mathcal{T}$)};
    \draw[solidarrow] (taskset.south) -- (avail.north);
    
    \end{tikzpicture}
}
\vspace{5 mm}
    \caption{(a) Column Generation (CG) decomposes an optimization problem into a master problem and a subproblem. The master problem selects the best combination of solutions from a solution pool ($\Omega$), while the subproblem uses dual values from the master to generate promising new candidates. Since $\Omega$ can be very large the CG operates on a smaller subset of solutions, $\textit{i.e.}$ $\Omega \subset \Omega'$. This process iterates until no improving solutions are found or a termination condition is met. (b) We propose Clustering-based Column Generation (CCG), which incorporates a clustering heuristic to efficiently solve the subproblem. To reduce computational cost, tasks in the orignal task set $\mathcal{T}$, are first grouped into clusters using $k$-means based on time windows. For each agent, a MIP with a time limit is solved on a reduced subset of tasks $\mathcal{T'}$, obtained by uniformly sampling from these clusters. The resulting solutions are added to the subset solution pool ($\Omega'$) for use in the master problem. 
    }
    \label{fig:CGnCCG}
\end{figure*}
\vspace{10 mm}

\par
\noindent \subsubsection{Restricted Master Problem}
This problem aims to identify the best set of solutions from the existing solution pool, resembling a set covering problem \citep{caprara2000algorithms}. Therefore, to solve the restricted master problem, we adopt a set covering problem formulation \citep{desrosiers2005primer}. A solution for the set covering problem selects a subset of sets (in our case, task sequences and modes for each agent) from a collection (the solution pool $\Omega'$) such that most of the elements (tasks) are covered. This must be done while adhering to constraints, such as each task being assigned to only one agent and maximizing the given objective. The set covering formulation-based restricted master problem finds optimal solutions from a subset of feasible candidate solutions $\Omega'$. Each candidate in $\Omega'$ is a feasible solution that specifies a sequence of addressing different tasks in the associated modes and arrival times. The pool $\Omega'$ can be initialized by some trivial solutions and can be updated iteratively with better solutions obtained by solving the subproblem.  The set covering formulation for our restricted master problem is given by:

\small
\begin{equation} \label{eq:rmp1}
\underset{\theta^{k}}{\max} \, \sum_{k \in \Omega'} c_k \cdot \theta^{k}
\end{equation}
\normalsize
\vspace{-0.1in}
\noindent subject to:
\small
\begin{equation} \label{eq:rmpc1}
\begin{split}
    \sum_{k \in \Omega'} a^{k}_i \cdot \theta^{k} \leq 1 \,\, \forall i \in \mathcal{T}/\{0\}, \\
\end{split}
\end{equation}
\begin{equation} \label{eq:rmpc1b}
\sum_{k \in \Omega'} b^{k}_{i,m} \cdot \theta^{k} \leq 1 \,\, \forall i \in \mathcal{T}/\{0\} , \forall m \in \mathcal{M}
_i\end{equation}
\vspace{-0.1in}
\begin{equation} \label{eq:rmpc2}
\sum_{k \in \Omega'}  \theta^{k} \leq |\mathcal{A}|,
\end{equation}
\vspace{-0.1in}
\begin{equation} \label{eq:rmpc3}
\theta^{k} \in \{0,1\}
\end{equation}
\normalsize

 \noindent Here, $\theta^k$ is a binary solution selection variable, where $\theta^k = 1$ if solution $k \in \Omega'$ is selected, and $\theta^k = 0$ otherwise. The objective function in Eq.~\eqref{eq:rmp1} selects the candidates from $\Omega'$ that maximize the original objective function Eq.~\eqref{eq:schedds_obj}. The constraint in Eq.~\eqref{eq:rmpc1} ensures that each task in $\mathcal{T}$ is assigned at most once. Here, $a^{k}_i$ is 1 if $i^{th}$ task is covered in $k^{th}$ solution. The constraint in Eq.~\eqref{eq:rmpc1b} ensures that mode $m$ for task $i$ is used at most once with $b^k_{i,m}$ being 1 if $i^{th}$ task in $k^{th}$ solution is done in mode $m$. The main purpose of Eq.~\eqref{eq:rmpc1b} is to identify the modes to explore via dual values. Finally, Eq.~\eqref{eq:rmpc2} ensures the number of candidates selected from $\Omega'$ does not exceed the number of available agents. \textcolor{black}{Typically, a linear relaxation of the original restricted master problem is solved in column generation~\citep{vanderbeck2005implementing}. This is because the dual of a mixed integer programming (MIP) problem is not well defined, which poses a challenge since dual variables are essential for solving the subproblem. In contrast, dual variables are well defined for linear programs~\citep{guzelsoy2007duality}. The linear relaxation of the integer program defined in Eqs.~\eqref{eq:rmp1}--\eqref{eq:rmpc3} is obtained by relaxing the integrality constraint in Eq.~\eqref{eq:rmpc3} and allowing $\theta^k \in [0,1]$. This relaxation also enables faster computation, as algorithms such as the simplex method~\citep{nabli2009overview} can efficiently solve large-scale linear programs. This is particularly advantageous when the restricted column set $\Omega'$ contains a large number of variables.}

\par
\subsubsection{Clustering-based subproblem} 
In CG, the subproblem is solved iteratively to populate $\Omega'$. Solving the subproblem efficiently with high-quality candidate solutions enhances the CG's exploration of the solution space and accelerates convergence to better solutions. However, as the number of tasks and agents increases, the expanded solution space slows down the search for promising candidates. To address this, we introduce a clustering method inspired by cluster-first, route-second heuristics commonly used in routing and scheduling~\citep{liu2023heuristics,elango2011balancing}. Our subproblem solves the MIP problem in Eq.~\eqref{eq:submip} for a single agent at a time, on a restricted set of tasks $\mathcal{T}'$. All tasks in $\mathcal{T}$ are clustered based on time windows. We use a $k$-means clustering~\citep{bishop2006mixture} approach to create $n_c$ clusters, with $n_c$ being a user-defined parameter. Then $\mathcal{T'}$ is generated by uniformly sampling each cluster.
These modifications allow CCG to converge to higher-quality solutions faster than the baseline, as demonstrated in Section~\ref{sec:exp}.
\par
The subproblem creates a solution for a single-agent setup with a restricted task pool $\mathcal{T}'$. The subproblem formulation is:
\begin{equation} \label{eq:submip}
\begin{split}
    \underset{x_{i\scriptsize{\xrightarrow{}}j}, y_{i,m}}{\max} \, \sum_{r=0}^{R}  \sum_{i \in \mathcal{T}'} \sum_{j \in \mathcal{T}'}  \sum_{m \in \mathcal{M}_i} \lambda_r \cdot f_{r}(x_{i\scriptsize{\xrightarrow{}}j}, y_{i,m}, s_i)  -  \sum_{i \in \mathcal{T}'/\{0\}} (\gamma_i \cdot \sum_{j \in \mathcal{T}'} x_{i\scriptsize{\xrightarrow{}}j}) - \sum_{i \in \mathcal{T}'/\{0\}} \sum_{m \in \mathcal{M}_i} (\beta_i \cdot y_{i,m}),
\end{split}
\end{equation}
\normalsize
\par
\noindent subject to constraints Eq.~\eqref{eq:scstr1}-\eqref{eq:scstr10b} and the number of agents is restricted to one. The objective function of the subproblem, Eq.~\eqref{eq:submip}, is a modified variant of Eq.~\eqref{eq:sched1}, tailored for a single-agent setup and augmented with dual variables from the restricted master problem. Here, $\gamma_i$ and $\beta_{i,m}$ are non-negative duals of the restricted master problem corresponding to constraints Eq.~\eqref{eq:rmpc1} and Eq.~\eqref{eq:rmpc1b} respectively. These dual variables are used to identify promising new candidate solutions. Dual variables can be directly obtained from the LP relaxation solution provided by the optimizer~\citep{laborie2018ibm}. It should be noted that the solution of the subproblem is added to $\Omega'$. 

\begin{algorithm}[t]
\footnotesize
    \caption{Clustering-based Column Generation (CCG) 
} 
    \label{algo:column_generation}
    \begin{algorithmic}[1]
        \STATE \textbf{Input:} Problem data ($\mathcal{A},\mathcal{T}, \mathcal{M}, T_H, Q$ and associated data for nodes), user preference $\lambda$, and algorithm parameters ($n_c,|\mathcal{T'}|,\text{max-iterations}, \text{time-limit}$). \label{lst:ln1}
        \vspace{-3mm}
        \STATE \textbf{Output:} Selected solution for each agent from $\Omega'$ \label{lst:ln2}
        
    \STATE $\mathcal{T}_{c}^{init} \gets$ Divide tasks in $\mathcal{T}$ to $n_c$ clusters based on time-windows using $k$-means clustering.\label{lst:ln3}
        \STATE $\Omega' \gets$ Initialize the candidate pool \label{lst:ln4}
        
        \WHILE{True} \label{lst:ln5}
            \STATE $\theta^k, \gamma_i, \beta_{i,m} \gets$ Solve Eq.~\eqref{eq:rmp1}-\eqref{eq:rmpc2} subject to $\theta^k \in [0,1]$ \label{lst:ln6}
            \STATE $\mathcal{T}_v \gets$ Initialize empty list to store visited customers. \label{lst:ln7}
            \FOR{$i = 1$ to $|\mathcal{A}|$} \label{lst:ln8}
                \IF{$i=1$} \label{lst:ln9}
                \STATE $\mathcal{T}_c \gets \mathcal{T}_{c}^{init}$ \label{lst:ln10}
                \ELSE \label{lst:ln11}
                    \STATE $\mathcal{T}_c \gets$ Create $n_c$ clusters of remaining tasks ($\mathcal{T}/\mathcal{T}_v$) using $k$-means clustering. \label{lst:ln12}
                \ENDIF \label{lst:ln13}
                \STATE $\mathcal{T'} \gets$ Uniformly sample tasks from $\mathcal{T}_c$ \label{lst:ln14}
                \STATE $\mathcal{S} \gets$ Solve the subproblem using Eq.~\eqref{eq:submip} \label{lst:ln15}
                
                \IF{$\mathcal{S}$ is not empty} \label{lst:ln16}
                    \STATE $\Omega' \gets$ Update solution pool with all feasible solutions from $\mathcal{S}$ \label{lst:ln17}
                    \STATE Update $\mathcal{T}_v$ with tasks addressed in the best solution of $\mathcal{S}$ \label{lst:ln18}
                \ENDIF \label{lst:ln19}
            \ENDFOR \label{lst:ln20}
            
            \IF{time-limit reached or $\text{max-iterations}$ completed} \label{lst:ln21}
                \STATE \textbf{Terminate} \label{lst:ln22}
            \ENDIF \label{lst:ln23}
        \ENDWHILE \label{lst:ln24}
        
        \IF{Terminate} \label{lst:ln25}
            \STATE Solve Eq.~\eqref{eq:rmp1}-\eqref{eq:rmpc2} subject to $\theta^{k} = \{0,1\}$ \label{lst:ln26}
        \ENDIF \label{lst:ln27}
        \STATE \textbf{return} Selected solution for each agent from $\Omega'$\label{lst:ln28}\end{algorithmic}
\end{algorithm}
\par
\subsubsection{Complete Algorithm} The CCG algorithm is detailed in Alg.~\ref{algo:column_generation}. In line~\ref{lst:ln1}, the algorithm receives problem parameters, user preferences, and key algorithm settings, including the number of clusters ($n_c$), the size of the restricted task set ($|\mathcal{T'}|$), maximum iterations, and the time limit. In line~\ref{lst:ln2}, initial task clusters $\mathcal{T}_{c}^{init}$ are formed by dividing $\mathcal{T}$ into $n_c$ clusters based on time windows using $k$-means. In line~\ref{lst:ln3}, the candidate solution pool $\Omega'$ is initialized, with initial solutions generated by randomly selecting tasks and modes while ensuring constraints in Eq.~\eqref{eq:scstr1}-\eqref{eq:scstr10b} are satisfied.
\noindent From line~\ref{lst:ln5}, the restricted master problem and subproblem are solved iteratively to generate new solutions. The restricted master problem is updated in each iteration based on $\Omega'$ (line~\ref{lst:ln6}), and the subproblem's objective is guided by the value of dual variables obtained from the restricted master. A list $\mathcal{T}_v$ is initialized to track visited tasks (line~\ref{lst:ln7}). The subproblem is solved iteratively for each agent in $\mathcal{A}$ (lines~\ref{lst:ln8}-\ref{lst:ln20}).
\noindent In the first iteration, task clusters $\mathcal{T}_c$ are initialized to $\mathcal{T}_{c}^{init}$ (line~\ref{lst:ln10}). For subsequent iterations, clusters are recreated from the remaining tasks (line~\ref{lst:ln12}). The restricted task set $\mathcal{T'}$ is formed by uniformly sampling clusters from $\mathcal{T}_c$ (line~\ref{lst:ln14}), after which the subproblem is solved (line~\ref{lst:ln15}). Feasible solutions with positive reduced costs are added to $\Omega'$ (line~\ref{lst:ln17}), and $\mathcal{T}_v$ is updated with tasks corresponding to the best solutions in $\mathcal{S}$ (line~\ref{lst:ln18}). This process continues until the stopping criteria—maximum iterations, time limit, or lack of solutions with positive reduced costs—are met (lines~\ref{lst:ln21}-\ref{lst:ln23}). Upon termination, the restricted master problem is solved to optimality with integrality constraints to obtain the final solution (lines~\ref{lst:ln25}-\ref{lst:ln28}). A time limit is applied to 
the restricted master and subproblems to maintain time efficiency during the process. 
\par
\textcolor{black}{The worst-case time complexity of CCG is determined by its components. The restricted master problem is a linear program, which, when solved using the simplex method, has a worst-case exponential complexity of $O(2^m)$  \citep{voulgaropoulou2019computational}, where $m$ is the number of constraints. However, it typically runs in polynomial time in practice \citep{voulgaropoulou2019computational}. The subproblem, solved as an MIP using a branch-and-bound solver, also possesses exponential complexity \citep{huang2021branch}. Additionally, the $k$-means clustering step has a worst-case complexity of $O(n^d)$, where $n$ is the number of nodes and $d$ is the data dimension \citep{arthur2006slow}. However, in practical implementations, $k$-means typically converges in $O(nkT)$ time \citep{arthur2006slow}, where $k$ is the number of clusters and $T$ is the number of iterations. With a fixed iteration limit, its runtime becomes effectively $O(1)$ per instance. Since both the master and subproblem have fixed time limits, their complexity per iteration is effectively $O(1)$. With an efficient implementation of $k$-means that also runs in bounded time, the overall time complexity is reduced to $O(I)$, where $I$ is the number of iterations until the convergence or time-out of the CCG. The space complexity depends on the size of the solution pool, leading to $O(|\Omega'|)$.}
\vspace{-2mm}
\subsection{M$^3$RS for Multi-Robot Disinfection} 
\textcolor{black}{We consider a fleet of surface disinfection robots deployed in hospital-like environments to provide around-the-clock disinfection. Specifically, we use G-robots \citep{mehta2022far}, which are mobile manipulators equipped with far-UVC emitters and are designed for surface disinfection (see Fig.~\ref{fig:grobot_hw}).} 
A disinfection mission is a collection of disinfection tasks spread in an environment such as a hospital. Each mission has a specified time horizon within which it should be completed. Further, the tasks must be done in their specified time window. The tasks can have different disinfection dose requirements. This dosage dictates disinfection quality, energy consumption, and the task service time. Each task can be disinfected in a particular set of modes, where the modes have different disinfection dosage options for that task. 
\par
The disinfection dosage ($D$ in $\frac{\text{Watt}\cdot s}{m^2}$) for UV-based surface disinfection is the product of UV exposure duration ($E_t$ in $s$) and irradiance of the UV light source ($I_r$ in $\frac{\text{Watt}}{m^2}$), given by $D = E_{t} \cdot I_{r}$~\citep{mehta2023uv}. \textcolor{black}{A higher disinfection dosage ($D$), requires a longer exposure time ($E_t$), given the fixed irradiance of the UV light source ($I_r$). Disinfection dosage is typically expressed as the percentage reduction in pathogen populations and is pre-calculated for various pathogen types~\citep{mehta2023uv}.} For example, a dosage of $D_{90}$ implies a $90\%$ reduction of the pathogenic population. A higher reduction percentage requires a stronger disinfection dose, which in turn leads to longer exposure times. Longer exposure times further imply higher energy consumption. We consider four disinfection modes that are $D_{99.9999}$, $D_{99.99}$, $D_{99}$, and $D_{90}$. Here, $D_{99.9999}$ is the strongest disinfection dose with the highest disinfection time and energy consumption as compared to $D_{90}$ \citep{mehta2023uv}. 
The set of modes can be determined by the acceptable disinfection levels for a surface. So, each task has an associated set of acceptable disinfectant doses or modes. 
The assignment of robots to different tasks and the corresponding modes will be determined by solving M$^3$RS. \textcolor{black}{The multi-objective formulation and execution mode decision variables in M$^3$RS enable trade-offs between disinfection quality and number of tasks completed, based on mission constraints and user preferences.}

For disinfection applications, we have two objectives:~i) the number of tasks completed and, i.e. $f_1$, ii) overall disinfection quality i.e. $f_2$. Higher disinfection quality requires more time and resources, limiting availability for other tasks. For example, critical locations like ICUs in hospitals require higher dosages, so higher quality is preferable. In contrast, areas like waiting rooms or offices do not need the highest quality, and covering the maximum number of surfaces might be preferable. Our formulation creates schedules by using user-specified trade-off requirements. \textcolor{black}{The overall objective of M$^3$RS--composed of $f_r(\cdot),~r = 1,2$, in Eq.~\eqref{eq:sched1}--for the disinfection application is given by:} 

\begin{equation} \label{eq:schedds_obj}
 \underset{x^{k}_{i{\xrightarrow{}}j}, y^{k}_{i,m}}{\max} \, \frac{1}{|\mathcal{T}|} \cdot \sum_{k \in \mathcal{A}} \sum_{i \in \mathcal{T}}  \sum_{j \in \mathcal{T}} \sum_{m \in \mathcal{M}_i} (\lambda)\cdot f_{1}+ (1-\lambda) \cdot f_{2} ,
\end{equation}
\normalsize
\normalsize
\noindent subject to the constraints in Eqs.~\eqref{eq:scstr1}--\eqref{eq:scstr10b}. This 
$f_1(\cdot)$ and $f_2(\cdot)$ are defined as follows:
\begin{equation}
f_{1} = x^{k}_{i \rightarrow j},
\end{equation}
\noindent which maximizes the quantity of tasks completed, and 
\begin{equation}
f_{2} = \frac{p_{i,m}}{p^{\max}} \cdot y^{k}_{i,m},
\end{equation}
\noindent which maximizes the quality of completed tasks. Here, $p_{i,m}$ denotes the reward representing the disinfection quality for task $i$ executed in mode $m$ (defined in Sec.~\ref{sec:syn-data}), $p^{\text{max}}$ is the maximum achievable disinfection quality reward, and $\lambda$ captures the user-defined preference. See Table~\ref{tab:nots} for a summary of the notation. When $\lambda = 1$, the objective prioritizes maximizing the number of tasks completed, disregarding disinfection quality. Conversely, when $\lambda = 0$, it exclusively focuses on maximizing disinfection quality. Users can adjust $\lambda$ based on mission priorities. Lower values suit missions with critical areas like ICUs or patient care rooms. Higher values are appropriate for missions with less critical spaces, e.g., waiting rooms. This mixed integer optimization model can be solved using a standard branch-and-cut algorithm. However, for larger problem instances, the branch-and-cut 
has an exponentially increasing computation time. The CCG algorithm, 
Alg.~\ref{algo:column_generation}, can be used 
in such cases.
\vspace{-2mm}
\section{Evaluation}\label{sec:exp}
\par 
In this section, the M$^3$RS problem and the proposed solution of CCG are evaluated. The evaluation is conducted for multi-robot disinfection scenarios. First, the metrics used for evaluation are defined. 
\textcolor{black}{Then, the impact of using task-level disinfection quality as a decision variable is analyzed. This is done by comparing M$^3$RS with a baseline that performs classic \textbf{r}outing and \textbf{s}cheduling in \textbf{f}ixed \text{(RS-F)} disinfection modes. Fixed modes are used since existing MRTA methods do not treat quality as a decision variable, and no relevant baselines exist. The comparison is performed on synthetically generated multi-robot disinfection missions.} Additioannly, the proposed CCG scheme is compared with a mixed integer programming (MIP) solver on synthetically generated missions. Finally, hardware and simulated case studies are presented.
\par
\textcolor{black}{In each evaluation, multiple problem instances are generated. Each instance is characterized by mission time, given by:} 

\small
\begin{equation}\label{eq:mts}
T_H = \frac{ t_{avg} \times \text{\# Tasks} }{\eta \times \text{Fleet size}}.
\end{equation}
\normalsize
\textcolor{black}{
Here, $T_H$ represents the mission time, \textit{i.e.}, the total time available to complete all tasks, while $\eta$ is the load factor used to simulate missions with varying $T_H$. \textcolor{black}{The load factor is commonly used in MRTA to represent mission workload relative to the robot's available capacity~\citep{botros2023optimizing}. We use the load factor to model varying mission durations, where a higher $\eta$ corresponds to tighter time constraints for the mission.} 
The average task completion time is given by $t_{avg}$. In our experiments, $t_{avg}$ is considered to be 4.5 minutes. We assume that \textcolor{black}{each} robot travels at a fixed speed of $0.5 \, {m}/{s}$. We use the MIP solver from IBM's CPLEX \citep{laborie2018ibm}. All evaluations are done on a machine with an Intel i9 processor and 32GB RAM.}
\vspace{-2mm}
\subsection{Evaluation Metrics}
We define 
\textcolor{black}{six} evaluation metrics to compare the performance of M$^3$RS 
\textcolor{black}{against other methods.} The metrics are: 
\small
\begin{equation}\label{eq:sr}
    SR \, = \, \frac{\sum_{k \in \mathcal{A}} \sum_{i \in \mathcal{T}}  \sum_{j \in \mathcal{T}} x^{k}_{i{\xrightarrow{}}j}}{|\mathcal{T}|},
\end{equation}

\begin{equation}\label{eq:dq}
    DQ \, = \, \frac{\sum_{k \in \mathcal{A}} \sum_{i \in \mathcal{T}}  \sum_{m \in \mathcal{M}_i} p_{i,m} \cdot y^{k}_{i,m}}{|\mathcal{T}| \cdot p^{max}},
\end{equation}

\begin{equation}\label{eq:msi}
    MSI \, = \, \frac{SR+DQ}{2},
\end{equation}
\begin{equation}\label{eq:atq}
    ATQ \, = \, \frac{DQ}{SR},
\end{equation}
\begin{equation}\label{eq:cbr}
    QBR \, = \, \frac{DQ - DQ^{*}}{(\bar{Q}_{exp}-\bar{Q}_{exp}^{*})+(\bar{M}_{exp}-\bar{M}_{exp}^{*})}.
\end{equation}

\normalsize
The \textbf{success rate ($SR$)} measures the number of tasks completed successfully (Eq.~\eqref{eq:sr}). The \textbf{dosage quality ($DQ$)} reflects the disinfection quality of completed tasks (Eq.~\eqref{eq:dq}). The \textbf{mean success index ($MSI$)} is the average of $SR$ and $DQ$ (Eq.~\eqref{eq:msi}). This metric scores a solution on combined performance on the number of tasks completed and disinfection quality. The \textbf{average task quality ($ATQ$)} measures the average disinfection quality received by a serviced task (Eq.~\eqref{eq:atq}). 

The \textbf{quality-benefit ratio ($QBR$)} compares a target solution with a baseline solution in terms of the gains made in disinfection quality relative to the resource usage (Eq.~\eqref{eq:cbr}). It is a ratio of the difference in disinfection quality to the difference in resource expenditure, measured against a baseline solution. For the target solution, $\bar{Q}_{exp}$ and $\bar{M}_{exp}$ represent the fleet's average energy expenditure and makespan. The RS-F solution serves as the baseline with $DQ^{*}$ as disinfection quality, $\bar{Q}_{exp}^{*}$ as average energy, and $\bar{M}_{exp}^{*}$ as makespan, where makespan is time taken by robot to complete its assigned task during the mission. 
\par
The objective values (Eq.~\eqref{eq:schedds_obj}) are also used as a metric. This is achieved by measuring the difference between the best-known bound on the objective value ($o^{*}$) and the current solution's objective value, $o$. MIP solvers typically provide the bound by solving a relaxed version of the problem. This metric is presented as the \textbf{relative optimality gap ($OG$)}~\citep{kaliszewski2021cooperative}: 


\begin{equation} \label{eq:rog}
    OG \, = \, \frac{o^{*} - o}{o^{*}} \cdot 100 \, \%.
\end{equation}
\normalsize
\par 


\begin{figure}[b!]
\noindent
\begin{minipage}[t]{0.48\textwidth}
\centering
\vspace{-28ex}
\begin{tabular}{|c|c|}
\hline
{Quantity} & {Value} \\
\hline
Layout area & $150, 200, 225 \, m^2$ \\
Number of tasks & $30-60$ \\
Proportion of critical tasks & $15-25\%$ \\
Avg. task completion time & $4.5 \, mins$ \\
Task surface area range & $0.5-2.2 \, m^2$ \\
$\eta$ & 1 \\
Fleet size & 4 \\
$Q$ & $6\text{Ah}$ \\
\hline
\end{tabular}
\captionof{table}{A list of parameters used to simulate disinfection missions to assess the impact of quality. A mission consists of disinfection tasks spread across a space, see Fig.~\ref{fig:layout}.}\label{tab:study1_params}
\end{minipage}
\hfill
\begin{minipage}[t]{0.48\textwidth}
\centering
\includegraphics[width=0.7\linewidth]{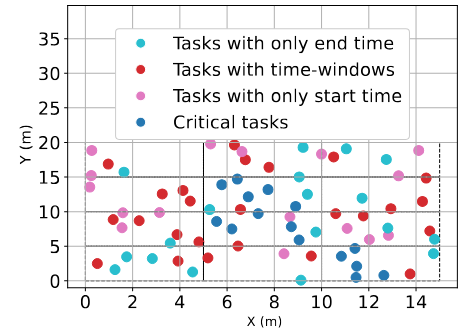}

\captionof{figure}{A sample physical layout of locations of different disinfection tasks. Here, the color coding is used to present different categories of tasks.}
\label{fig:layout}
\end{minipage}
\end{figure}
\subsection{M$^3$RS vs RS-F (\textcolor{black}{Routing and Scheduling with Fixed modes})}\label{sec:exp1}
\par
\textcolor{black}{In this experiment, we assess the effectiveness of our proposed M$^3$RS formulation in multi-robot disinfection scenarios. The results demonstrate that M$^3$RS outperforms baseline approaches across key performance metrics.}
\subsubsection{Baselines} 
\textcolor{black}{We evaluate the impact of the M$^3$RS formulation, which integrates quality levels as a decision variable through multiple execution modes, against RS-F, a routing and scheduling approach with fixed execution modes. RS-F follows the vehicle routing problem with time windows~\citep{kallehauge2005vehicle} and does not account for task-level quality. We consider two RS-F variants: i) RS-F (Min), which defaults to the lowest disinfection dosage, and ii) RS-F (Max), which defaults to the highest disinfection dosage.} For M$^3$RS we use different values of user preference \textit{i.e.} $\lambda$. A higher $\lambda$ prioritizes quantity over quality. Synthetically generated data is used to benchmark the problems, as detailed in the following paragraph.  Both M$^3$RS and RS-F (Min and Max) are solved using an MIP solver with a 10-minute time limit per problem instance. The evaluation metrics include $SR$, $DQ$, $MSI$ $ATQ$, and $QBR$, see Eq.~\eqref{eq:sr}-\eqref{eq:cbr}. RS-F (Min) is used as the baseline for $QBR$ calculation since it operates in the lowest disinfection mode, allowing us to measure the improvement in disinfection quality relative to resource expenditure. We analyze the solutions based on working and idle time and also discuss the impact of relaxing task time windows for M$^3$RS.

\subsubsection{Synthetic Data Generation} \label{sec:syn-data}
The synthetic data simulates realistic hospital missions with rectangular layouts of varying sizes. \textcolor{black}{For each layout type, $10$ random problem instances are generated and solved.} 
A sample layout is shown in Fig.~\ref{fig:layout}. 
\textcolor{black}{We consider four disinfection modes, indexed by $m$, where a lower $m$ indicates higher disinfection quality. Specifically, $m = 0$ corresponds to $D_{99.9999}$ (highest quality), followed by $D_{99.99}$ ($m = 1$), $D_{99}$ ($m = 2$), and $D_{90}$ ($m = 3$, lowest quality). For task $i$, the mode set ($\mathcal{M}_i$) is randomly assigned between one and four disinfection modes. The associated quality reward (see Eq.~\eqref{eq:schedds_obj}) for each mode is defined as $p_{i,m} = 2^{-m}$  for $m \in \{0,1,2,3\}$. For example, selecting the highest quality mode ($m = 0$) for task $i$  yields the maximum reward of $1$, while the lowest quality mode ($m = 3$) yields a reward of $0.125$. 
For each disinfection task, the target surface is assumed to be planar, with an area randomly assigned between $0.5$ and $2.2 \, m^2$. The required dosage, exposure time, and energy consumption for disinfecting the surface are computed based on UV disinfection dosage and energy models~\citep{mehta2022far}.
} 
\textcolor{black}{A summary of the experimental setup is provided in Table~\ref{tab:study1_params}.} Each problem instance involves a fleet of four robots. We set the value of load factor $\eta = 1$ for this experiment. This ensures that for the disinfection mission there is sufficient time to address the tasks. The mission time ($T_H$) is calculated using Eq.~\eqref{eq:mts}. Tasks are categorized as critical or non-critical. 
In each problem instance, a subset of critical tasks, representing $15-25\%$ of the total tasks, must be completed within a specified time window using the highest disinfection dosage, $D_{99.9999}$. 
These are crucial surfaces, such as those in patient rooms and ICUs. The inequality in Eq.~\eqref{eq:scstr1} is replaced by an equality constraint for critical tasks. Non-critical tasks, like those in waiting areas, washrooms, and office spaces, can be skipped but contribute to overall hygiene if addressed. These tasks may have fixed time windows, and deadlines, or be available throughout the mission. All time windows are uniformly sampled from the mission time horizon. 


\begin{figure*}[b!]
    \centering
    
   {\includegraphics[width=0.9\columnwidth]{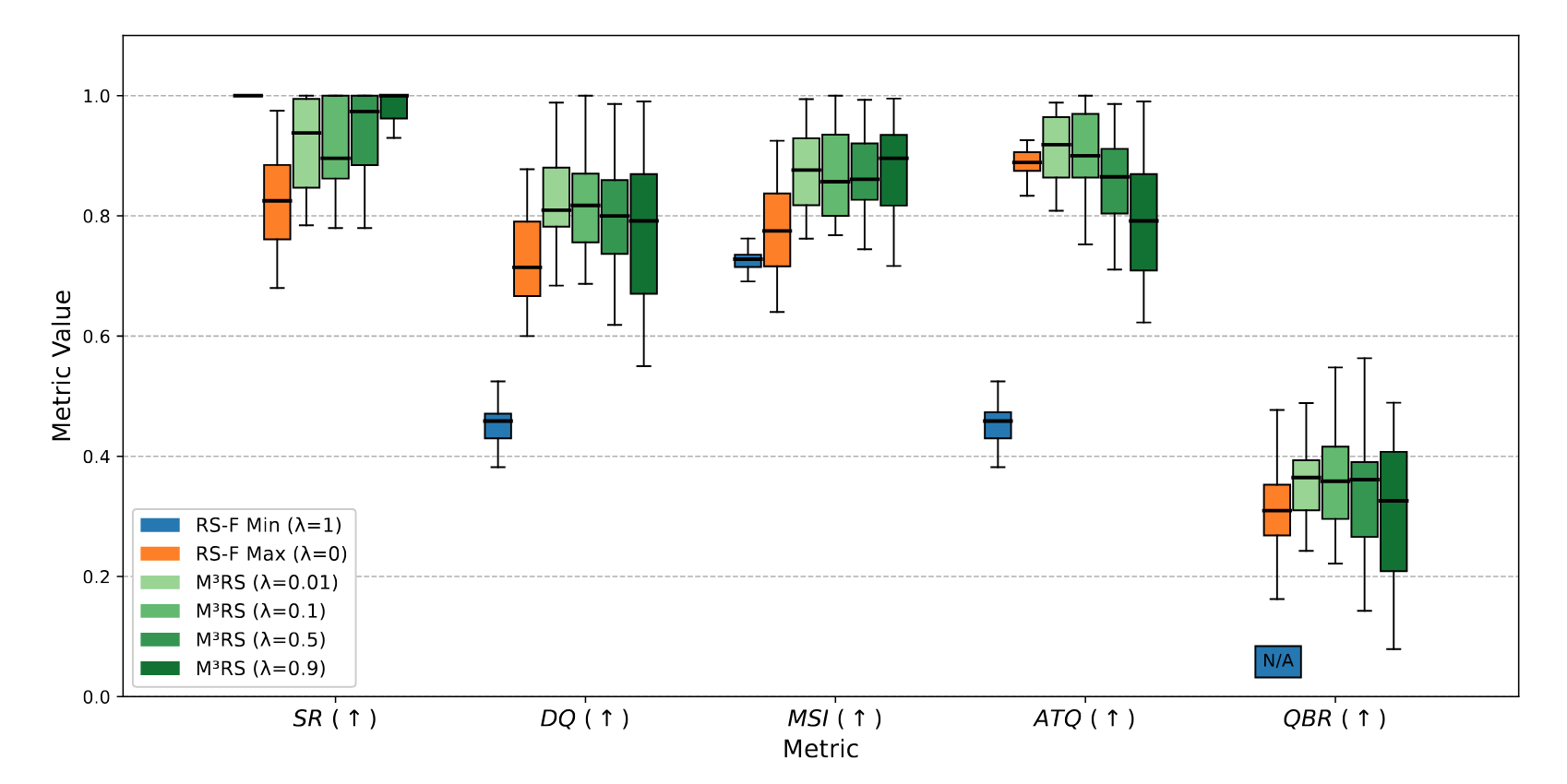}}

    \caption{Comparison of M$^3$RS and RS-F (Min/Max) solutions to assess the impact of quality of service (QoS) decision variables. Unlike RS-F, which considers only quantity, M$^3$RS explicitly incorporates QoS decision variables. Different $\lambda$ values in M$^3$RS prioritize quantity over quality. Both formulations are solved using  MIP-solver~\citep{laborie2018ibm} with a 10-minute time limit on synthetic instances (30–60 tasks, four robots). The figures depict box-plot generated using solutions of 15 random instances, with higher values of metrics indicating better performance. M$^3$RS achieves 1.5$\%$–45$\%$ gains across quality and quantity metrics.}
    \label{fig:m3rsVrsf}
\end{figure*}

\par 
\subsubsection{Results} 
\textcolor{black}{Fig.~\ref{fig:m3rsVrsf} presents the box-plots of performance metrics for RS-F and M$^3$RS variants}. 
RS-F (Min) uses the lowest disinfection mode. This minimizes execution time and maximizes success rate ($SR$), due to reduced task completion time. However, it results in the lowest disinfection quality ($DQ$). RS-F (Max) defaults to the highest disinfection mode. This increases execution time and prioritizes disinfection quality. However, it lowers the success rate due to longer disinfection times. M$^3$RS ($\lambda =0.01$) improves disinfection quality by 10\% over RS-F (Max) and 38\% over RS-F (Min) while maintaining a strong balance between efficiency and quality. In terms of the combined metric ($MSI$), M$^3$RS ($\lambda =0.01$) achieves 10\% and 15\% gains over RS-F (Max) and RS-F (Min), respectively, demonstrating its ability to optimize both success rate and disinfection quality. M$^3$RS solutions also achieve higher Average Task Quality ($ATQ$), with a 3\% gain over RS-F (Max) and 46\% over RS-F (Min). The Quality-Benefit Ratio ($QBR$) improves by 6\% compared to RS-F (Max), indicating that M$^3$RS provides better disinfection quality relative to resource expenditure (energy and time). These results indicate that incorporating variable disinfection modes yields better overall disinfection quality while maintaining or improving task completion rates. \textcolor{black}{To further ascertain the gains made by task execution modes, we conducted a statistical comparison of success rate $SR$ and disinfection quality $DQ$ between RS-F (Max) and M$^3$RS ($\lambda = 0.01$) using Welch’s t-test~\citep{sakai2018t}. This test was applied to evaluate whether the differences observed in each metric between the two variants are statistically significant. For the $SR$, the p-value was $7.74 \times 10^{-6}$, and for the $DQ$, the p-value was $5.08 \times 10^{-6}$. Both results indicate statistically significant differences at the $95\%$ confidence level, demonstrating that M$^3$RS ($\lambda = 0.01$) achieves significant improvements over RS-F (Max) in these metrics.
}
\begin{figure*}[h!]
    \centering
    
    \subfloat[Box-plot comparing idling time to $T_H$ ratio for M$^3$RS and RS-F for different number of tasks and a fleet size of 4.]{\includegraphics[width=0.4\columnwidth]{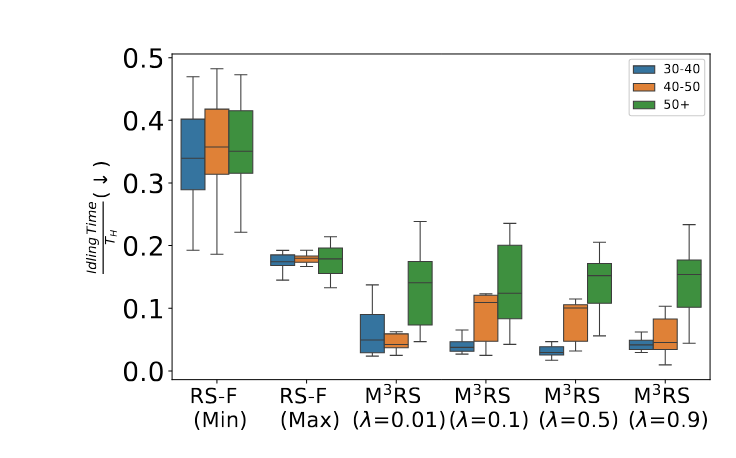}}
    \hspace{0.02\textwidth}
    \subfloat[Box-plot comparing working time to $T_H$ ratio for M$^3$RS and RS-F for different number of tasks and a fleet size of 4.]{\includegraphics[width=0.44\columnwidth]{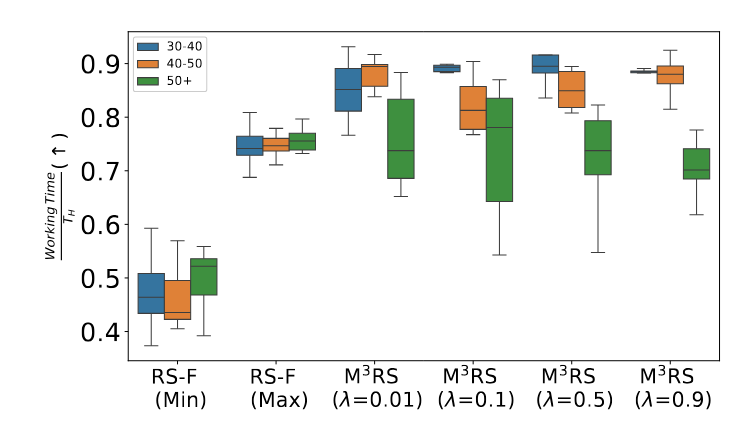}}
    \vskip\baselineskip
    \subfloat[Box-plot depicting impact on $SR$ and $DQ$ for different types of time windows. The tasks with time-windows could be critical or non-critical.]{\includegraphics[width=0.4\columnwidth]{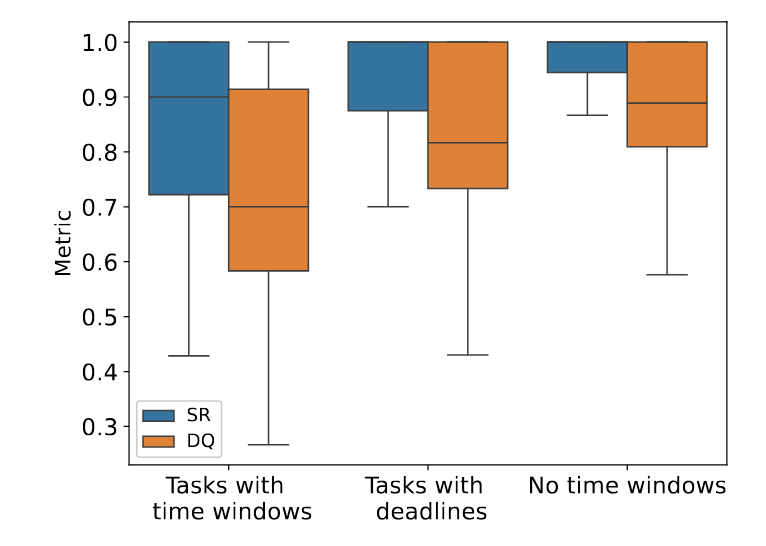}}
   
    \caption{(a) \& (b) The solution of M$^3$RS makes effective use of agents as indicated by lower idling times and higher working times. (c) Further, tasks available during the entire mission have a higher task completion rate and disinfection quality.}
    \label{fig:study1_tw}
\end{figure*}
\par 
We further analyze the missions by looking at the ratio of idle time to task servicing time, as well as the $SR$ and $DQ$ for different time-window types (see Fig.~\ref{fig:study1_tw} (a)-(c)). The solution of M$^3$RS shows less idle time and more working time, indicating better use of mission time; see Fig.~\ref{fig:study1_tw} (a)-(b).  
\textcolor{black}{Use of variable disinfection modes promotes better use of time and energy, which could be spent on disinfecting more tasks or improving quality for others, depending on the user preference \textit{i.e.} $\lambda$}. In most cases, RS-F (Max) shows higher idle time and lower working time than M$^3$RS, as disinfection in the highest modes requires robots to arrive at or before a task’s time window. Early arrivals lead to waiting, which limits task completion and results in lower SR and DQ compared to M$^3$RS. \textcolor{black}{For missions with more than 50 tasks, we observe that for certain $\lambda$ values, M$^3$RS shows slightly higher average idling times and lower average working times compared to RS-F (Max). This \textcolor{black}{could be attributed} to the increased problem size with more tasks and may be improved with more efficient solution strategies.}

\par
When examining $SR$ and $DQ$ across time-window types, tasks available throughout the mission have the best performance, while those with limited time windows perform worse, see Fig.~\ref{fig:study1_tw} (c). This suggests that relaxing time constraints, where possible, could lead to better overall performance.
\par
\textcolor{black}{Overall, the synthetic experiments demonstrate that incorporating quality as a decision variable in M$^3$RS enables more effective planning of disinfection missions, balancing the number of tasks completed with the quality of service within given time and resource constraints. While this work focuses on disinfection, M$^3$RS is also applicable to other domains, such as cleaning robots, where balancing cleaning efficiency and time is critical~\citep{megalingam2025cleaning}, and robot patrolling, which involves managing trade-offs between high- and low-priority areas~\citep{katole2023balancing}.}

\subsection{Performance of CCG}\label{sec:cgvsmip}

\textcolor{black}{We evaluate our clustering-based column generation (CCG) method for solving M$^3$RS by comparing it with 
a standard MIP solver~\citep{laborie2018ibm}. The results show that CCG achieves comparable or better performance in less time.} 
\par 
\subsubsection{Baselines} The performance of solving M$^3$RS with CCG is compared with that of directly solving using an MIP solver. For CCG, we report results with varying number of clusters i.e $K \in [2,4,6,8]$. The performance is evaluated in terms of the relative optimality gap ($OG$) given by Eq.~\eqref{eq:rog}. For CCG, we use the best bound from the MIP solver to compute the $OG$.

\subsubsection{Setup} 
\begin{figure*}[h!]
    \centering
    
    \subfloat[$\eta=1.0$ and $\lambda = 0.9$]{\includegraphics[width=0.45\columnwidth]{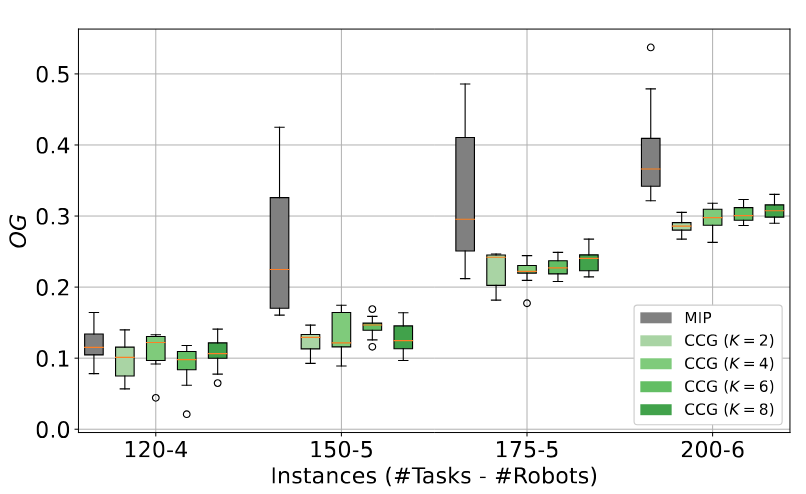}}
    \hspace{0.02\textwidth}
    \subfloat[$\eta=2.0$ and $\lambda = 0.9$]{\includegraphics[width=0.45\columnwidth]{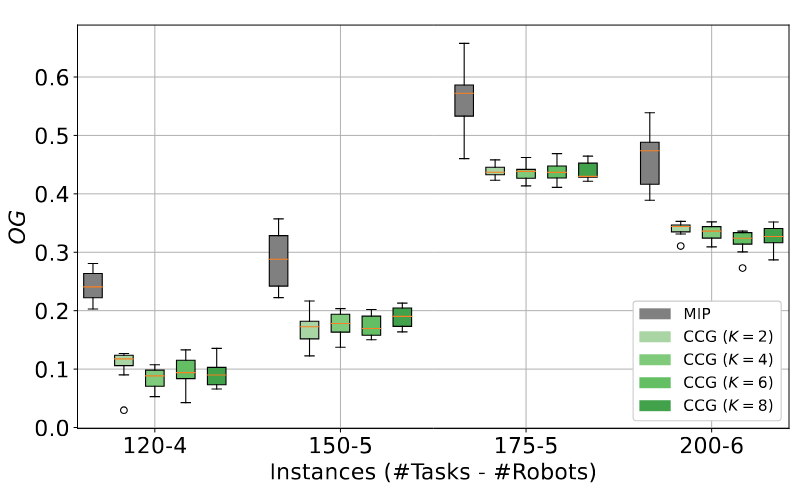}}
    \vskip\baselineskip
    \subfloat[$\eta=1.0$ and $\lambda = 0.1$]{\includegraphics[width=0.45\columnwidth]{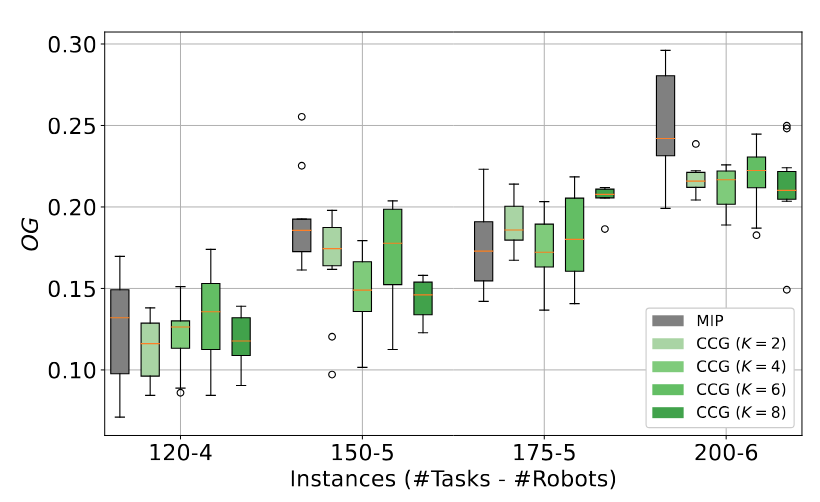}}
    \hspace{0.02\textwidth}
    \subfloat[$\eta=2.0$ and $\lambda = 0.1$]{\includegraphics[width=0.45\columnwidth]{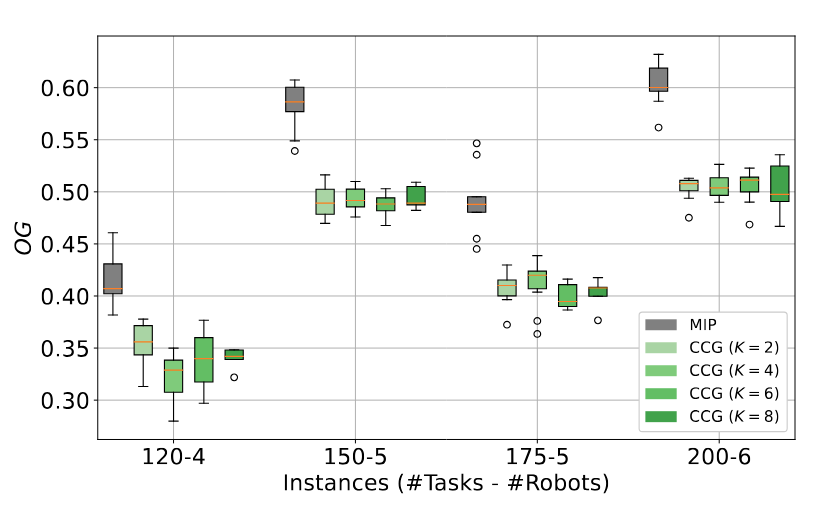}}
   
    \caption{(a)-(d) \textcolor{black}{Box-plots comparing the optimality gap ($OG$) of M$^3$RS problems solved by: (i) a standard MIP solver~\citep{laborie2018ibm}, and (ii) our Clustering-based Column Generation (CCG) approach.} CCG is evaluated with different numbers of clusters ($K = 2, 4, 6, 8$). Both methods are given a 10-minute time limit. Box-plots are generated from multiple randomly generated problem instances, varying in $\lambda$ and $\eta$. Lower $OG$ values indicate better performance. CCG consistently achieves a lower $OG$ than the MIP solver within the same time budget.}
    \label{fig:study2_ccgvsmip}
\end{figure*}

For this, four problem types are considered with a number of tasks and robots mentioned in Fig.~\ref{fig:study2_ccgvsmip}. For each problem type, 10 instances are randomly generated using the procedure outlined in Sec.~\ref{sec:exp}-B. User preference for each instance is sampled from $\lambda \in [0.1,0.9]$, with higher $\lambda$ values prioritizing task completion over disinfection quality. We use load factors $\eta = 1$ and $\eta = 2$, where a higher $\eta$ simulates missions with shorter $T_H$, providing less time for mission completion. The mission time ($T_H$) is calculated using Eq.~\eqref{eq:mts}. The time limit for the MIP solver and CCG is set to 10 minutes. We further report the performance with a time limit for MIP set to 25 minutes.

\begin{table}[b]
\centering
\caption{Average optimality gap ($OG$) difference between CCG and the MIP solver across various instances, with a 25-minute time limit for MIP and a 10-minute time limit for CCG. Positive values indicate better $OG$ for MIP, while negative values favor CCG. All values are given in $\%$. Colors highlight \ho{MIP outperforming CCG} and \hy{CCG outperforming MIP}. Our CCG method achieves comparable or better $OG$ while using  60\% less computation time than the MIP solver.}
\label{tab:cgvsmiptime}
{%
\begin{tabular}{|lcc|cccc|}
\hline
\multirow{2}{*}{Metric}& \multirow{2}{*}{$\eta$} & \multirow{2}{*}{$\lambda$} & \multicolumn{4}{|c|}{\# tasks - \# robots.} \\
\cline{4-7}
 & & & $120-4$ & $150-5$ & $175-5$ & $200-6$ \\
\hline
\multirow{4}{*}{$OG_{CCG} -OG_{MIP}$} & \multirow{2}{*}{$1.0$} & $0.9$ & \ho{$4.1\%$} & \hy{$-7.2\%$} & \hy{$-3.2\%$} & \hy{$-6.0\%$} \\ 
                        &                         & $0.1$ & \ho{$5.0\%$} & \hy{$-3.0\%$} & \ho{$0.9\%$} & \ho{$0.6\%$} \\
                        & \multirow{2}{*}{$2.0$} & $0.9$ & \ho{$0.7\%$} & \hy{$-15.7\%$} & \hy{$-4.2\%$} & \hy{$-15.4\%$} \\
                        &                         & $0.1$ & \hy{$-0.1\%$} & \hy{$-8.6\%$} & \hy{$-6.2\%$} & \hy{$-9.2\%$} \\
\hline
\end{tabular}}
\end{table}
\par
\subsubsection{Discussion} \textcolor{black}{The $OG$ metric for MIP and CCG is presented as box-plots in Fig.~\ref{fig:study2_ccgvsmip} (a)-(d)}.
The $OG$ achieved by CCG solutions is consistently better than that of the MIP solver with the same time limit, see Fig.~\ref{fig:study2_ccgvsmip} (a)-(d). 
The change in $OG$ for the MIP solver relative to CCG, with the time limit for the MIP solver set to 25 minutes, is reported in Table~\ref{tab:cgvsmiptime}. 
\textcolor{black}{Even with an increased time limit of 25 minutes for the MIP solver, CCG achieves comparable or better performance while using only 10 minutes of computation time, representing a 60$\%$ reduction.} This indicates that clustering and sampling approach coupled with column generation provides an effective way to find good solutions in a shorter time as compared to the MIP solver. It is evident from the results that clustering improves the overall solution optimality. The results indicate that tuning the number of clusters is non-trivial. \textcolor{black}{While CCG achieves a better $OG$ than the MIP solver, the $OG$ tends to increase with instance size. Nonetheless, clustering remains a promising strategy for improving scalability and solution quality.} 
Integrating learning-based heuristics for clustering and formulating CCG using learning-based column generation methods \citep{yuan2024reinforcement} could further enhance performance.

\subsection{Simulated and Hardware Case Study}

\textcolor{black}{In this section, simulation and hardware case studies are presented which apply M$^3$RS in realistic scenarios.} 
Videos of the experiments are available on the project page: {\href{https://sites.google.com/view/g-robot/m3rs/}{https://sites.google.com/view/g-robot/m3rs/}}.

\subsubsection{Simulation and Hardware Experiment Setup} 
\begin{figure*}[b!]
    \centering
    \subfloat[A team of G-robots disinfecting tables.]{\includegraphics[width=0.35\textwidth]{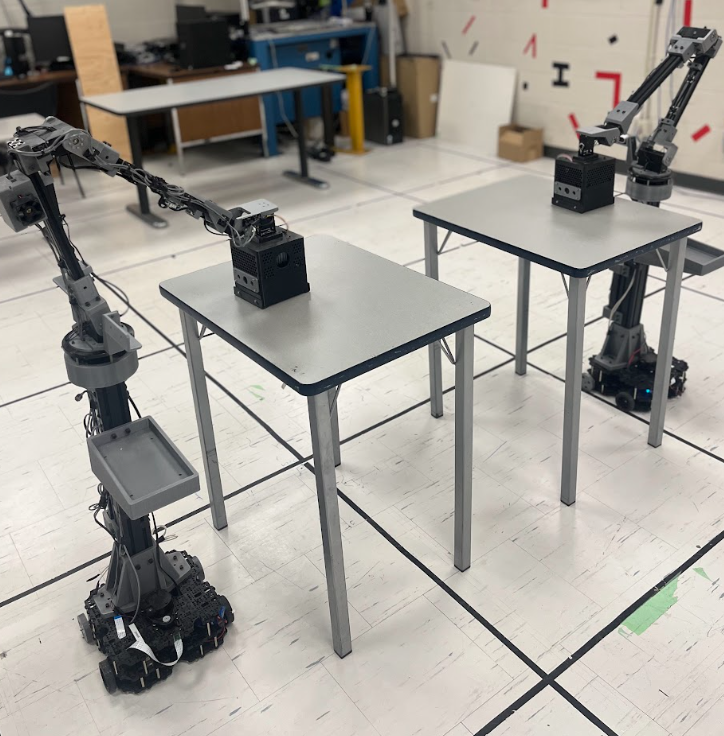}\label{fig:grobot_hw}}
    \hfill
    \subfloat[The simulated environment with 20 tasks and 2 robots.]{\includegraphics[width=0.55\textwidth]{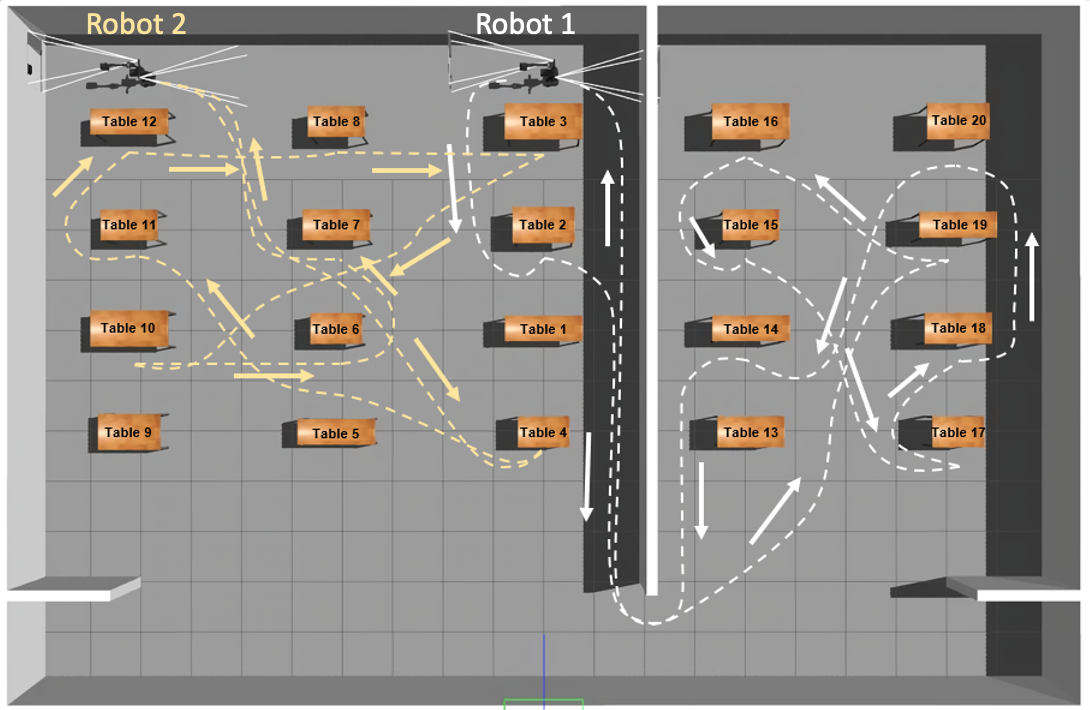}\label{fig:gazebo_rt}}
    \vspace{-0.2in}
    \vskip\baselineskip
    \subfloat[The route sequence for hardware experiment with $\lambda = 0.9$ ]{\includegraphics[width=0.25\textwidth]{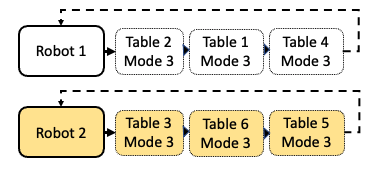}}
     \hfill
     \subfloat[The route sequence for hardware experiment with $\lambda = 0.1$ ]
     {\includegraphics[width=0.2\textwidth]{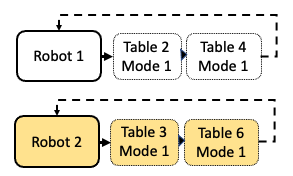}}
     \hfill
     \subfloat[The route sequences for simulated mission with 20 tasks, 2 robots, $T_H$ of 0.4 hours and $\lambda=0.9$.  ]{\includegraphics[width=0.52\textwidth]{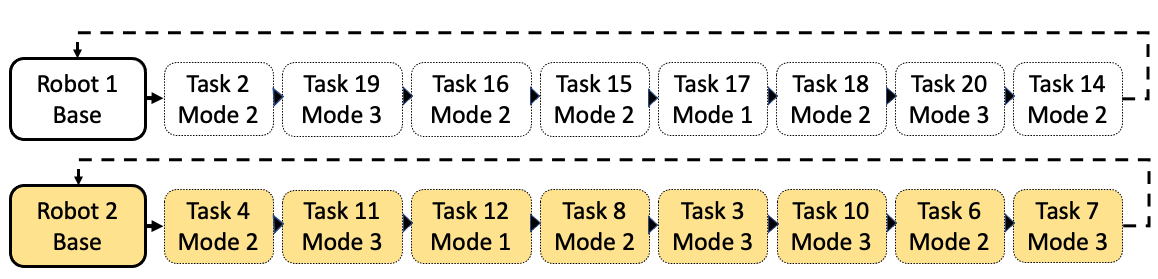}\label{fig:rts_0.8}}

    \caption{ (a) A team of G-robots~\citep{mehta2022far} disinfecting tables. (b) The routes of two robots executing various tasks in the simulated environment. The route of the first robot is highlighted in white and for the second robot yellow is used. (c) \& (d) The route sequence taken by each robot for the hardware mission with $\lambda = 0.9$ and $0.1$. For $\lambda = 0.9$, the emphasis is on disinfecting all the tables at higher disinfection modes. Whereas, for $\lambda = 0.1$, the emphasis is on ensuring high-quality disinfection rather than disinfecting all tables. (e) The route sequence taken by each robot in the simulated mission. For $\lambda = 0.9$, maximization of the number of tasks completed has more emphasis than the quality of disinfection. As a result, most of the tasks are done in higher modes. }
    \label{fig:sim_exp}
    \vspace{-0.1in}
\end{figure*}

Multiple disinfection mission instances are generated for a simulated environment. Each disinfection mission is characterized by mission time, time windows for each task, and the number of robots used in the mission. The simulated environment consists of $20$ tables as disinfection surfaces spread across two rooms. The fleet size considered in the range, $|\mathcal{A}| = 2$ to $5$. The mission time is calculated using Eq.~\eqref{eq:mts}. The entire simulated experiment is conducted in ROS \citep{ros} and the Gazebo environment \citep{koenig2013many}. A simulated mission is depicted in Fig~\ref{fig:sim_exp} (b).
\par
For the hardware case study, we use two G-robots (shown in Fig.~\ref{fig:sim_exp} (a)) for a disinfection mission consisting of six tables. Each table can be disinfected in three disinfection modes, i.e., $D_{99.99}$, $D_{99}$, and $D_{90}$. The entire disinfection mission must be completed in $17$ minutes.
\noindent For both experiments, M$^3$RS is solved for two preferences: $\lambda \, = 0.9$ and $0.1$ for each mission. The optimization model is solved in CPLEX~\citep{laborie2018ibm}. The reported metrics are $SR$, $DQ$, $MSI$, and the average lateness of the robot in task completion.
\par
\subsubsection{Discussion} The metrics for simulated and hardware case studies are reported in Table~\ref{tab:sim_exp}. The example of generated route sequences in a simulated experiment is visualized in Fig.~\ref{fig:sim_exp} (b) \& (e), respectively. The route sequence for hardware experiments is shown in Fig.~\ref{fig:sim_exp} (c)-(d). From Table~\ref{tab:sim_exp}, it can be seen that a decent $SR$ is achieved in all missions. Even for missions with tight $T_H$, the $SR$ is high. Therefore, a larger robotic fleet would be necessary to compensate for tighter mission times to service as many tasks as possible. For missions with shorter $T_H$, there is a decline in the $DQ$ metric. This is because with tighter mission times, each task has a smaller time window, which means many of the tasks cannot be performed in higher disinfection modes. A possible remedy could be to enable robots to collaborate on different tasks. \textcolor{black}{Finally, delays in task completion times were observed in both simulation and hardware implementations. These are primarily due to longer-than-expected navigation times caused by environmental and planning uncertainties. Such delays can be mitigated by incorporating chance constraints to model system stochasticity more effectively.}

\renewcommand{\tabcolsep}{5pt}
\begin{table}[t!]
    \centering
    \caption{Results for simulated and hardware experiments. }
    \begin{tabular}{|ccc|cccc|}
    \hline
          \# tasks - \# robots & $T_H$ (hrs) & $\lambda$ & $SR$ & $DQ$ & MSI  &  Avg. Lateness (s) \\
         \hline
        \multirow{2}{*}{20-2} & \multirow{2}{*}{0.40} & 0.9 & 0.80 & 0.57 & 0.63  & 1.9   \\
         & & 0.1 & 0.70 & 0.71 & 0.60  & 6.2  \\
         \hline
         \multirow{2}{*}{20-3} & \multirow{2}{*}{0.21} & 0.9 & 0.80 & 0.51 & 0.60  & 30.1   \\
         & & 0.1  & 0.75 & 0.56 & 0.58  & 1.4  \\
         \hline
         \multirow{2}{*}{20-4} & \multirow{2}{*}{0.20} & 0.9 & 0.95 & 0.55 & 0.74  & 33.6   \\
         & & 0.1 & 0.90 & 0.60 & 0.72  & 19.1  \\
         \hline
        \multirow{2}{*}{20-5} & \multirow{2}{*}{0.16}  & 0.9 & 0.80 & 0.57 & 0.68  & 29.4   \\
         & & 0.1 & 0.75 & 0.62 & 0.68  & 33.7  \\ 
         \hline
         \multirow{2}{*}{6-2*} & \multirow{2}{*}{0.28} & 0.9 & 1.00 & 0.40 & 0.70  & 27.0   \\
         & & 0.1 & 0.67 & 1.00 & 0.65  & 32.0  \\
        \hline
        \multicolumn{6}{l}{\scriptsize{ * = Results for hardware experiments.}}
    \end{tabular}
    \label{tab:sim_exp}
\vspace{-0.2in}
\end{table}

\section{Conclusions and Future work} \label{sec:conc}

In this work, we proposed M$^3$RS, a novel multi-objective formulation that routes and schedules a team of agents to accomplish various tasks in a mission. The agents can perform the task in one of many available modes. Each mode has distinct task quality, resource consumption, and service times. We first propose a mixed-integer linear programming model for M$^3$RS. \textcolor{black}{We showcase the utility of M$^3$RS in multi-robot disinfection applications.  The experiments suggest that our M$^3$RS demonstrates 3\%–46\% performance improvements over standard task allocation methods across various metrics. The solutions for M$^3$RS provide more flexibility to users to select solutions based on desired preferences. We also presented a clustering-based column generation approach that solves M$^3$RS instances in $60\%$ lower compute time while maintaining competitive solutions.} 

The current approach 
assumes a deterministic environment and \textcolor{black}{simplified robot energy models}. 
Our future work will extend M$^3$RS to incorporate elements like stochastic travel times, dynamic arrival of tasks, \textcolor{black}{realistic robot energy models}, and realistic issues like robot breakdown. Further, we aim to tightly integrate the robot path planning with the decision-making framework provided by M$^3$RS. Finally, we will investigate using deep reinforcement learning techniques to generate good-quality solutions in real-time. 

\vspace{-0.12in}
\section*{Acknowledgments}
\par 
We would like to thank Mr. Matthew Lisondra for all the support with the experiments.
\vspace{-0.15in}
\bibliographystyle{apalike}
\renewcommand{\bibfont}{\small}
\bibliography{bib}

\end{document}